\documentclass[preprint,review,3p]{elsarticle}

\usepackage{amssymb}
\usepackage{amsmath}

\newcommand{\EQ}{\begin{equation}}
\newcommand{\EN}{\end{equation}}
\newcommand{\EQA}{\begin{eqnarray}}
\newcommand{\ENA}{\end{eqnarray}}

\usepackage{mathrsfs}
\usepackage{hyperref}
\usepackage{graphicx}
\usepackage{makecell}
\usepackage{multirow}
\biboptions{sort&compress}
\usepackage{algorithm}
\usepackage{algorithmic}
\usepackage{tcolorbox}
\usepackage{listings}
\usepackage{booktabs,siunitx,makecell}
\usepackage{tabularx}
\usepackage{array}
\definecolor{commentgreen}{RGB}{2,112,10}
\definecolor{eminence}{RGB}{108,48,130}
\definecolor{weborange}{RGB}{255,165,0}
\definecolor{frenchplum}{RGB}{129,20,83}

\journal{Computers and Fluids}

\begin{document}

\begin{frontmatter}



\title{RocketAgent: A Long-Horizon Engineering Agent for Multidisciplinary Design of Liquid-Rocket Thrust Chambers}


\author[a,b]{Junxiang He}
\author[a,c,d]{Runze Mao\corref{author}}
\author[c]{Kun He}
\author[c]{Teng Zhang}
\author[c]{Liming Zheng}
\author[c]{Ke Xiao}
\author[a,c,d]{Zhi X. Chen\corref{author}}

\cortext[author]{Corresponding authors.\\
\textit{E-mail addresses:} maorz1998@pku.edu.cn (R. Mao), chenzhi@pku.edu.cn (Z. X. Chen).}
\affiliation[a]{organization={State Key Laboratory of Turbulence and Complex Systems, School of Mechanics and Engineering Science, \\Peking University},
            city={Beijing},
            postcode={100871}, 
            country={China}}

\affiliation[b]{organization={College of Engineering, Peking University},
            city={Beijing},
            postcode={100871}, 
            country={China}}

\affiliation[c]{
    organization={Legendspace Intelligence Technology Co., Ltd.}, 
    city={Beijing},
    postcode={100080},
    country={China}
}

\affiliation[d]{organization={AI for Science Institute (AISI)},
            city={Beijing},
            postcode={100084}, 
            country={China}}


\begin{abstract}
Liquid-rocket thrust-chamber design involves interdependent analyses in which downstream constraints can require earlier design decisions to be revisited. Managing these dependencies across heterogeneous tools requires consistent design information and coordinated updates throughout the workflow. We present RocketAgent, a long-horizon engineering agent for multidisciplinary preliminary design of liquid-rocket thrust chambers. A single plan-owning Coding Agent coordinates engineering skills for performance sizing, subsystem optimization, geometry generation, and multiphysics assessment. A provenance-aware knowledge graph supports method selection, while a typed Design Intermediate Representation maintains shared parameters, artifacts, and decisions. Revision-aware checks invalidate affected results and block superseded inputs, with consequential changes subject to engineering approval. In a representative simulation-based design, RocketAgent continued from an infeasible cooling search through an engineer-authorized operating-point revision, identified feasible subsystem designs, and coordinated subsequent geometry generation and multiphysics assessment to support final configuration selection. Separate module tests assessed surrogate predictions and nozzle adaptation. A two-configuration comparison across three controlled scenarios verified the expected dependency invalidations and superseded-input blocking before solver execution. The representative case demonstrates sustained coordination across a multidisciplinary design workflow, while the controlled tests establish the behavior of the revision mechanisms supporting that execution.
\end{abstract}




\end{frontmatter}




\newpage
\section{Introduction}
\label{sec:introduction}
The design of complex engineering products involves interdependent decisions across disciplines and computational tools. Requirements, model assumptions, and analysis results must remain consistent as a design develops, while findings from later stages can require earlier decisions to be revised \cite{hirshorn2017nasa,singh2018engineering}. Managing these dependencies requires engineers to determine which tasks can proceed, which results remain applicable, and which analyses must be repeated after a change. Supporting this coordination is an engineering informatics problem that connects the representation of design knowledge with its use in an evolving workflow.

Liquid-rocket thrust-chamber design provides a concrete setting for studying this problem. Propulsion performance, regenerative cooling, injector design, and structural response impose coupled requirements on the same design \cite{huzel1992modern,cohn2010liquid,kim2014multidisciplinary}. An operating point selected for high specific impulse may prove incompatible with cooling constraints, prompting a revision that changes propellant flows, subsystem designs, and inputs to subsequent analyses. These changes must be carried consistently through specialized calculation, geometry, and simulation tools. Completing the preliminary design therefore requires coordination across successive engineering decisions, including feedback that redirects work to an earlier stage \cite{huzel1992modern,hirshorn2017nasa}.

Established engineering approaches provide foundations for managing this process. Knowledge-based engineering (KBE) and knowledge graphs (KGs) formalize reusable knowledge and its relationships, supporting method selection and design decisions \cite{verhagen2012kbe,hao2021decision}. Digital threads connect design information across stages \cite{singh2018engineering}, while multidisciplinary design optimization organizes coupled analyses around explicit variables, objectives, and constraints \cite{martins2013multidisciplinary}. These approaches address complementary aspects of knowledge representation, information continuity, and computational design. Their integration into an executable workflow requires the represented knowledge and dependencies to inform what work is performed and how its results affect subsequent decisions.

Large language model (LLM)-based agents offer an approach to this coordination through knowledge retrieval, planning, and tool use \cite{lewis2020rag,yao2023react,liu2026idesigngpt}. Engineering applications have connected agent decisions to deterministic calculations for code-compliant structural design, finite-element optimization, and traceable analysis workflows \cite{chen2025rcmultiagent,chen2026uhpcmultiagent,qin2026structureclaw}. Such systems provide precedents for using computational feedback to guide further actions. However, LLM outputs can contain unsupported content \cite{ji2023hallucination}, making engineering knowledge, explicit interfaces, and numerical checks essential to their application. For an extended multidisciplinary workflow, these controls must operate across task boundaries: a locally valid calculation may still use an outdated operating point, depend on an unaccepted subsystem design, or produce results that become inapplicable after a revision.

The problem addressed here is how to connect agent planning with the evolving engineering basis of a design. The agent needs access to applicable methods, accepted inputs, unresolved constraints, and the dependencies between computational results. When an assessment fails, the workflow must distinguish input errors, numerical failures, and violations of engineering constraints, while keeping an unsuccessful bounded search distinct from a proof of physical infeasibility. When a change is accepted, affected results must be identified before they are reused. These requirements motivate the integration of reusable engineering knowledge, persistent design state, and revision handling within the execution process.

We present RocketAgent, a long-horizon engineering agent for multidisciplinary preliminary design of liquid-rocket thrust chambers. Here, \emph{long-horizon execution} denotes continued progression through an extended sequence of dependent engineering tasks, including feedback-driven revisions and expert decision points. A single Coding Agent maintains the global plan and coordinates engineering skills through declared inputs, outputs, and validation requirements. A provenance-aware KG supports method selection, while a typed Design Intermediate Representation (Design IR) links requirements, parameters, artifacts, evidence, and decisions. Accepted changes trigger dependency-scoped invalidation and an ordered re-execution plan; solver-entry checks block superseded inputs. Deterministic tools perform calculations and enforce numerical criteria, while engineers authorize consequential changes and review the final design.

We contribute an architecture connecting these mechanisms and evaluate their operation at complementary levels. A representative simulation-based case traces the workflow from accepted requirements through cooling-driven redesign, subsystem development, geometry generation, and multiphysics assessment. This record documents how the Agent carries work across stages and continues after engineering feedback. Separate module tests assess surrogate predictions and geometry-interface reuse. A two-configuration comparison across three controlled scenarios tests dependency invalidation and superseded-input blocking, with execution stopping before solver launch. Within the scope of simulation-based preliminary design, these evaluations together examine sustained workflow execution and the specific revision mechanisms that support it.

Section~\ref{sec:related-work} reviews related work, Section~\ref{sec:methodology} presents the framework, and Section~\ref{sec:evaluation-results} reports the engineering and controlled-test results. Sections~\ref{sec:discussion} and \ref{sec:conclusions} discuss their implications and summarize the contributions.

\section{Background and Related Work}
\label{sec:related-work}

\subsection{ Knowledge-Based, Knowledge-Graph, and Model-Based Engineering Design}
Knowledge-based engineering (KBE) formalizes product and process knowledge to automate recurring engineering tasks. Verhagen et al.~\cite{verhagen2012kbe} consolidated its foundations and identified challenges in traceability, interoperability, and implementation. Al-Ashaab et al.~\cite{alashaab2012kbeproval} connected automotive design changes to validation plans through KBE-ProVal, linking physical tests, theoretical knowledge, and historical results within a product-lifecycle-management environment. These approaches establish knowledge reuse and change traceability as practical design capabilities.

Knowledge graphs extend these capabilities to decision support and design exploration. Hao et al.~\cite{hao2021decision} structured decision knowledge for retrieval and navigation, using natural-language processing to transform compromise Decision Support Problem formulations into graph representations. Their evaluation covered 25 supply-chain decision problems. Liu et al.~\cite{liu2022kgdesign} used graph embeddings to predict functional-coupling relations in smart products. Huang et al.~\cite{huang2025kgarchitecture} combined a multi-source KG with graph matching, multi-objective optimization, decision making, and mechanism-based simulation for launch-vehicle stage-separation architectures. Their method makes the decision uses of graph content explicit: component constraints screen architecture combinations, decision knowledge supplies optimization criteria and methods, and mechanism nodes connect selected configurations to simulation models.

Model-based systems engineering and digital threads supply complementary information-management principles. Madni and Sievers~\cite{madni2018mbse} described an evolving system model supporting specification, design, validation, and configuration management. Singh and Willcox~\cite{singh2018engineering} linked lifecycle information through stagewise updates for iterative design under uncertainty. Together, these foundations motivate a separation between reusable engineering knowledge and the evolving state of a particular design. RocketAgent implements this separation through a provenance-aware KG and a shared Design IR, then connects both to executable engineering skills.

\subsection{ LLM-Based Engineering Agents and Skill/Tool Use}
LLM-based workflows connect language interpretation to knowledge retrieval and executable actions. Retrieval-augmented generation conditions responses on external documents \cite{lewis2020rag}, while ReAct interleaves reasoning, actions, and observations \cite{yao2023react}. Engineering applications adapt these patterns to product development: DesignGPT organizes role-based collaboration with human designers \cite{ding2023designgpt}, natural-language CAD workflows connect requirements to calculations and editable geometry \cite{deng2024llmcad}, and iDesignGPT combines agent clusters, design methods, and a knowledge base for concept development and evaluation \cite{liu2026idesigngpt}. The CAD study demonstrated parameter-consistent generation of a one-stage reduction-gear system. iDesignGPT evaluated conceptual-design support through design-space metrics, expert assessment across six public challenges, and controlled user studies.

Physics-backed agents extend this interaction into computational design loops. Chen and Bao~\cite{chen2025rcmultiagent} used specialized agents and deterministic modules for code-compliant reinforced-concrete beam design. Their subsequent framework used LLM agents to formulate optimization inputs, refine the search from performance feedback, and support conversational design review, with genetic optimization and finite-element analysis providing the numerical evaluations for ultra-high-performance concrete beams \cite{chen2026uhpcmultiagent}. StructureClaw connects governed skills, typed tools, shared artifacts, and local analysis backends, evaluating execution traces across 150 controlled structural-engineering scenarios \cite{qin2026structureclaw}. Its benchmark covers standard workflows, interactive robustness, and multimodal structural-model reconstruction. These studies demonstrate executable feedback and increasingly explicit workflow governance for bounded engineering tasks. They also establish useful precedents for separating agent planning from deterministic calculation and for evaluating the resulting execution records.

For a thrust chamber, orchestration extends across thermochemistry, cooling, geometry, reacting flow, and structural assessment. RocketAgent focuses on driving this long task sequence through one plan-owning Coding Agent and contract-bound skills. A revision-aware Design IR carries inputs and artifacts between tasks, while source-linked execution and deterministic checks address the risk of unsupported LLM output \cite{ji2023hallucination}.

\subsection{ Computational Methods for Thrust-Chamber Design}
Thrust-chamber analysis combines specialized models with different inputs and assumptions. NASA CEA computes chemical-equilibrium compositions, thermodynamic properties, and theoretical rocket performance \cite{gordon1994cea}; RocketCEA exposes these calculations through Python \cite{taylor2026rocketcea}. RPA supports liquid-rocket performance analysis and thermal analysis of cooled chambers and nozzles \cite{ponomarenko2010rpa,ponomarenko2012thermal}. Their outputs inform geometry and cooling design, provided that units, equilibrium assumptions, and nozzle conventions remain aligned.

Established geometry and design methods include Rao's optimum-thrust nozzle contour \cite{rao1958nozzle}, the parametric thrust-chamber model of Rezende et al.~\cite{rezende2016parametric}, liquid-propellant engine design procedures \cite{huzel1992modern}, and fluid-cooled chamber criteria \cite{nasa1972fluidcooled}. PicoGK supplies programmable voxel, mesh, and lattice operations for computational geometry \cite{leap71picogk}. Coupled analyses resolve interactions within such geometries: Kim et al.~\cite{kim2014multidisciplinary} linked compressible reacting-flow calculations to regenerative cooling and compared performance predictions with hot-firing measurements, while Liu et al.~\cite{liu2020thermostructural} transferred thermal and pressure fields into a combustion-chamber structural model. These studies establish the physical links that a cross-tool design workflow must preserve during redesign.

Optimization methods search the resulting trade spaces. Martins and Lambe~\cite{martins2013multidisciplinary} organized multidisciplinary-design-optimization architectures around unified formulations and process flows. NSGA-II provides elitist multi-objective search \cite{deb2002nsgaii}, while efficient global optimization and surrogate-based methods reduce reliance on expensive function evaluations \cite{jones1998ego,forrester2009surrogate}. These methods assume that variables, objectives, constraints, and analysis interfaces have been defined. In requirement-driven design, this formulation must remain consistent as operating conditions and geometry evolve. RocketAgent uses a GA for regenerative-cooling-channel search and NSGA-II for surrogate-assisted centrifugal-injector screening, with each search connected to the shared design state.

Higher-fidelity methods provide flow-field evidence and numerical-credibility procedures. DeepFlame extends the OpenFOAM finite-volume framework for reacting-flow computation through Cantera, PyTorch, and additional computational utilities \cite{mao2023deepflame}. Assessments of OpenFOAM and EBIdnsFoam examine numerical accuracy and parallel performance \cite{zirwes2023assessment}. The VOF method resolves free-boundary dynamics \cite{hirt1981vof}. Discretization-uncertainty estimation \cite{celik2008procedure}, verification and validation principles \cite{oberkampf2002vv}, ASME V\&V 20 \cite{asme2009vv20}, and NASA-STD-7009B \cite{nasa2024models} guide interpretation of computational evidence. These methods supply the numerical capabilities and credibility practices on which the integrated workflow builds.

\subsection{Summary of Research Gap}
\label{subsec:research-gaps}
Existing research provides mature design methods and task-bounded closed loops in which computational evaluation guides design revision \cite{huang2025kgarchitecture,chen2025rcmultiagent,chen2026uhpcmultiagent,qin2026structureclaw}. The remaining gap concerns their joint execution for a complex industrial product such as a liquid-rocket thrust chamber. Its preliminary design requires knowledge, mutable parameters, heterogeneous models, and revision feedback to remain connected from requirements to the expert-reviewed design package. The reviewed digital-thread, artifact-centered agent, and model-credibility approaches provide complementary foundations \cite{singh2018engineering,qin2026structureclaw,nasa2024models}, but do not jointly address this coupled thrust-chamber workflow.

Driving this extended workflow requires three supporting capabilities: \emph{state continuity}, linking requirements and parameters to their sources and artifacts; \emph{evidence-closed execution}, connecting tool results to design checks and selection decisions; and \emph{feedback and governance continuity}, identifying affected dependencies and coordinating re-execution after changes. These capabilities allow the Agent to carry work across stage boundaries and continue after a design revision. For example, a mixture-ratio change affects thermal loads, cooling feasibility, and downstream geometry, requiring the plan and dependent results to be updated together. These challenges motivate an Agent-driven workflow that connects multidisciplinary tasks from requirements to design evaluation, supported by consistent information transfer and revision handling.

\section{Proposed Methodology and System Realization}
\label{sec:methodology}

RocketAgent drives thrust-chamber design through three connected stages: translating requirements into a design concept, developing its coupled subsystems, and verifying the resulting configurations. The following sections explain the engineering decisions at each stage and the shared-state mechanism that supports continuation when a design must change.

\subsection{Problem Formulation}
\label{subsec:problem-state}
The problem concerns steady, simulation-supported preliminary design of hundred-newton-class liquid-rocket thrust chambers. Given thrust and operating requirements, the design process must determine a consistent performance state, chamber/nozzle contour, regenerative-cooling layout, and preliminary swirl-injector configuration, then assess their integrated multiphysics response. Propellants, wall materials, nozzle families and angles, and subsystem configurations are replaceable inputs supported by the corresponding backend models.

The design space combines architectural choices with continuous operating and geometric parameters. Hard constraints enforce geometric consistency, thermal-hydraulic limits, structural margins, and model applicability; soft objectives compare performance and distance from active constraints. Some quantities remain open to optimization, and their eventual values become accepted design decisions. Because these decisions couple successive analyses, completing the task requires coordination across engineering knowledge, numerical tools, and design revisions.

\begin{figure*}[!t]
\centering
\includegraphics[width=\textwidth]{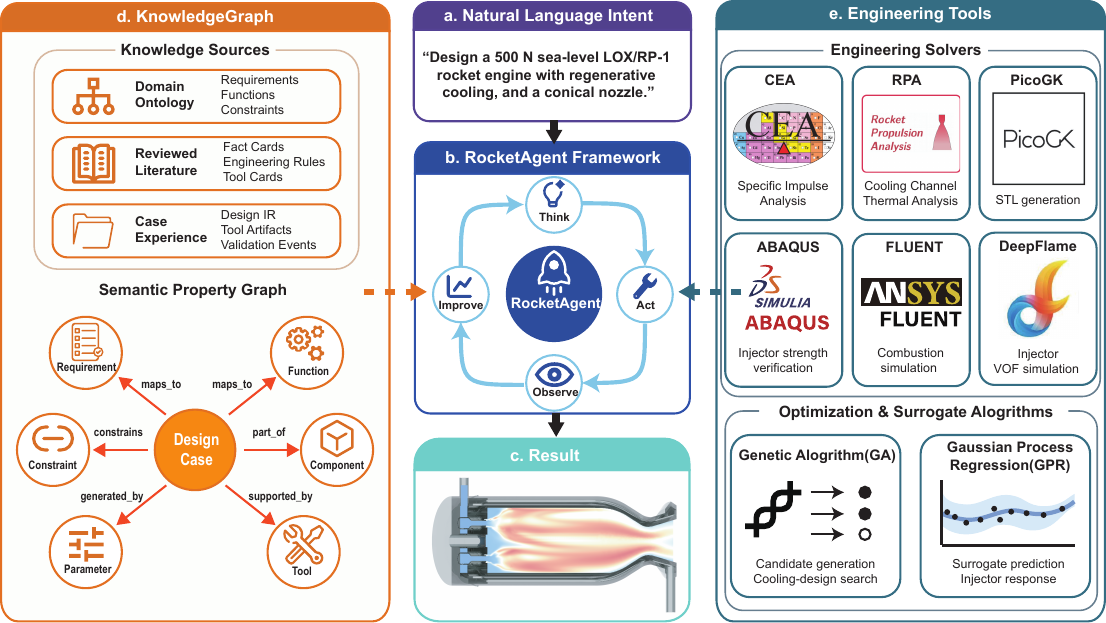}
\caption{RocketAgent architecture: (a) natural-language design intent; (b) the plan-owning Coding Agent and its Think--Act--Observe--Improve cycle; (c) an illustrative thrust-chamber representation; (d) the engineering KG; and (e) specialist calculation, geometry, simulation, and optimization capabilities. }
\label{fig:rocketagent-framework}
\end{figure*}

Fig.~\ref{fig:rocketagent-framework} translates this coordination problem into an architecture. Its central sequence, (a)--(c), connects natural-language intent to a preliminary-design output through a plan-owning Coding Agent. Within the cycle in (b), the Agent interprets the current task, invokes an engineering skill, examines the returned result, and updates the plan. The two supporting branches provide different inputs to this cycle: the KG in (d) supplies applicable knowledge and prior experience, while the calculation, geometry, simulation, and optimization tools in (e) evaluate the evolving design. The architecture therefore links reasoning about the next task to execution of that task. Section~\ref{subsec:overall-framework} develops this component-level view into the extended engineering workflow.

\subsection{Design-State Governance and Execution Architecture}
\label{subsec:overall-framework}
\label{subsec:state-governance}
\begin{figure*}[!t]
\centering
\includegraphics[width=\textwidth]{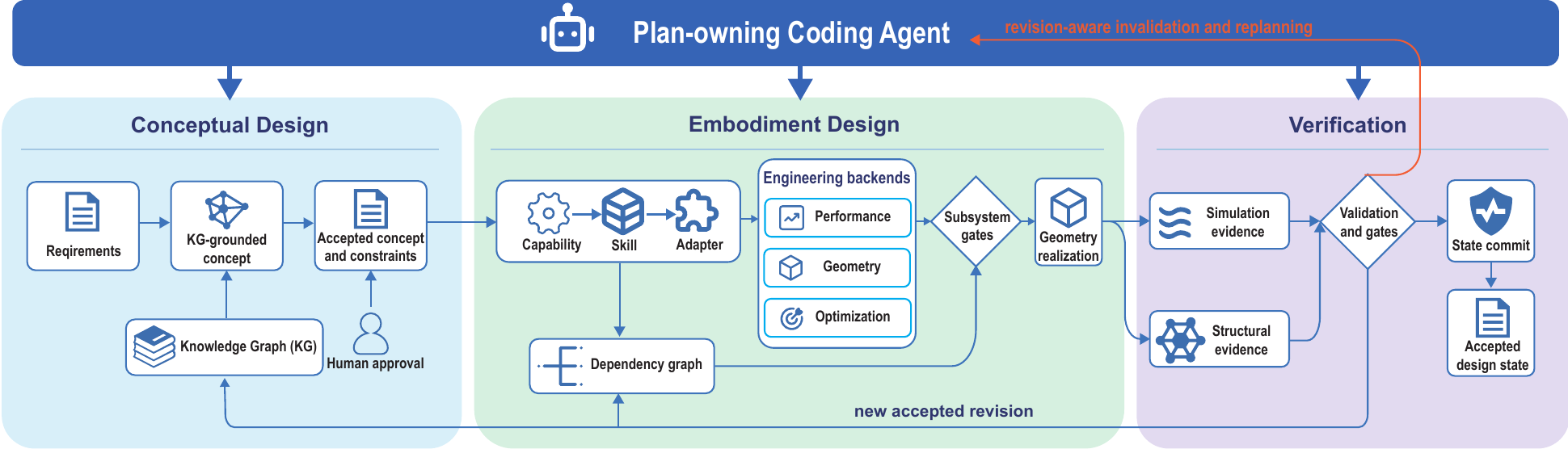}
\caption{Conceptual workflow of RocketAgent across Conceptual Design, Embodiment Design, and Verification, coordinated by a plan-owning Coding Agent with revision-aware feedback.}
\label{fig:rocketagent-workflow}
\end{figure*}

Fig.~\ref{fig:rocketagent-workflow} organizes the workflow into Conceptual Design, Embodiment Design, and Verification. The plan-owning Coding Agent spans all three stages: it maintains the task plan, selects eligible skills, inspects their outputs, and determines whether to advance or revisit earlier work. Long-horizon execution means driving this extended sequence from requirements to an accepted design state, including continuation after engineering feedback.

Conceptual Design combines requirements with reusable KG knowledge and human approval to establish the accepted concept and constraints (Section~\ref{subsec:kg-conceptual-design}). Embodiment Design translates this basis into performance, geometry, and optimization tasks through the capability--skill--adapter interface. The dependency graph identifies the prerequisites for these tasks, and subsystem gates check the resulting designs before geometry realization (Section~\ref{subsec:embodiment-design}).

Verification follows a fixed sequence: injector VOF, Fluent combustion and flow-distribution assessment, and finally Abaqus structural checking (Section~\ref{subsec:verification-refinement}). A candidate advances only after passing the current stage; failure at any stage rejects it and terminates its remaining verification tasks. The orange return in Fig.~\ref{fig:rocketagent-workflow} carries feedback into the broader design process, where an approved design-state change invokes revision-aware invalidation and Agent replanning. It does not permit a rejected candidate to continue through verification. The lower blue return represents the accepted revision carried into subsequent work. Knowledge reuse remains subject to the provenance and review requirements in Section~\ref{subsec:kg-conceptual-design}.

These cross-stage handoffs use the Design IR as the active preliminary-design digital thread \cite{singh2018engineering}. Whereas the KG supplies reusable methods and experience, the Design IR links the current requirements, parameters, artifacts, and decisions through units, provenance, dependencies, and revisions. For workflow step $k$, the accepted design snapshot and its checked update are
\begin{equation}
S_k=\langle Q_k,X_k,A_k,E_k,D_k\rangle, \qquad
S_{k+1}=\mathcal{C}_i\!\left[\mathcal{T}_i(S_k)\right],
\label{eq:state-transition}
\end{equation}
where $Q_k$ denotes accepted requirements and assumptions, $X_k$ accepted typed parameters, $A_k$ active artifact references, $E_k$ the evidence references supporting that accepted snapshot, and $D_k$ its committed design decisions. A skill transformation $\mathcal{T}_i$ proposes a successor, and $\mathcal{C}_i$ checks its inputs, engineering criteria, and required approval before commitment. A failed check leaves this accepted snapshot unchanged: proposed parameters, artifacts, evidence bindings, and design decisions do not replace $Q_k$, $X_k$, $A_k$, $E_k$, or $D_k$. Execution history is separate. Writing $L_k$ for the accumulated execution and evidence records, each attempt appends a record $\ell_i$ containing its input revision, output references, check results, feedback, and disposition, so $L_{k+1}=L_k\mathbin{\Vert}\ell_i$ even when the proposal fails. Rejection and cancellation of the candidate's remaining verification tasks are recorded in this execution layer; retention of the accepted snapshot does not authorize further verification of a rejected candidate. Acceptance permits downstream use, feasibility denotes satisfaction of the applicable hard constraints, and selection chooses among feasible alternatives.

Revision-aware automatic invalidation is integrated into the main state-update path. An accepted change identifies direct consumers, propagates through downstream dependencies, marks affected results stale, and generates an ordered re-execution plan. Unaffected outputs retain their valid bindings. Solver-entry checks additionally compare submitted artifacts with the active registry, blocking superseded files even when their existence and checksums are valid. Separate consumers for structural-only material fields permit local revisions without invalidating unrelated cooling or geometry results. Section~\ref{subsubsec:revision-aware-validation} evaluates these paths.

The capability--skill--adapter separation connects planning to execution: capabilities identify engineering tasks, skills define procedures and checks, and adapters translate the accepted state into backend inputs. Deterministic programs calculate responses and apply numerical gates; externally coordinated Fluent runs enter through the same evidence interface. Human approval governs consequential changes and final selection. Record and interface details are retained in ~\ref{app:implementation-details}, and the Agent's runtime and execution records in ~\ref{app:agent-execution}.

\subsection{KG-Grounded Conceptual Design}
\label{subsec:kg-conceptual-design}
\begin{figure*}[!t]
\centering
\includegraphics[width=0.98\textwidth]{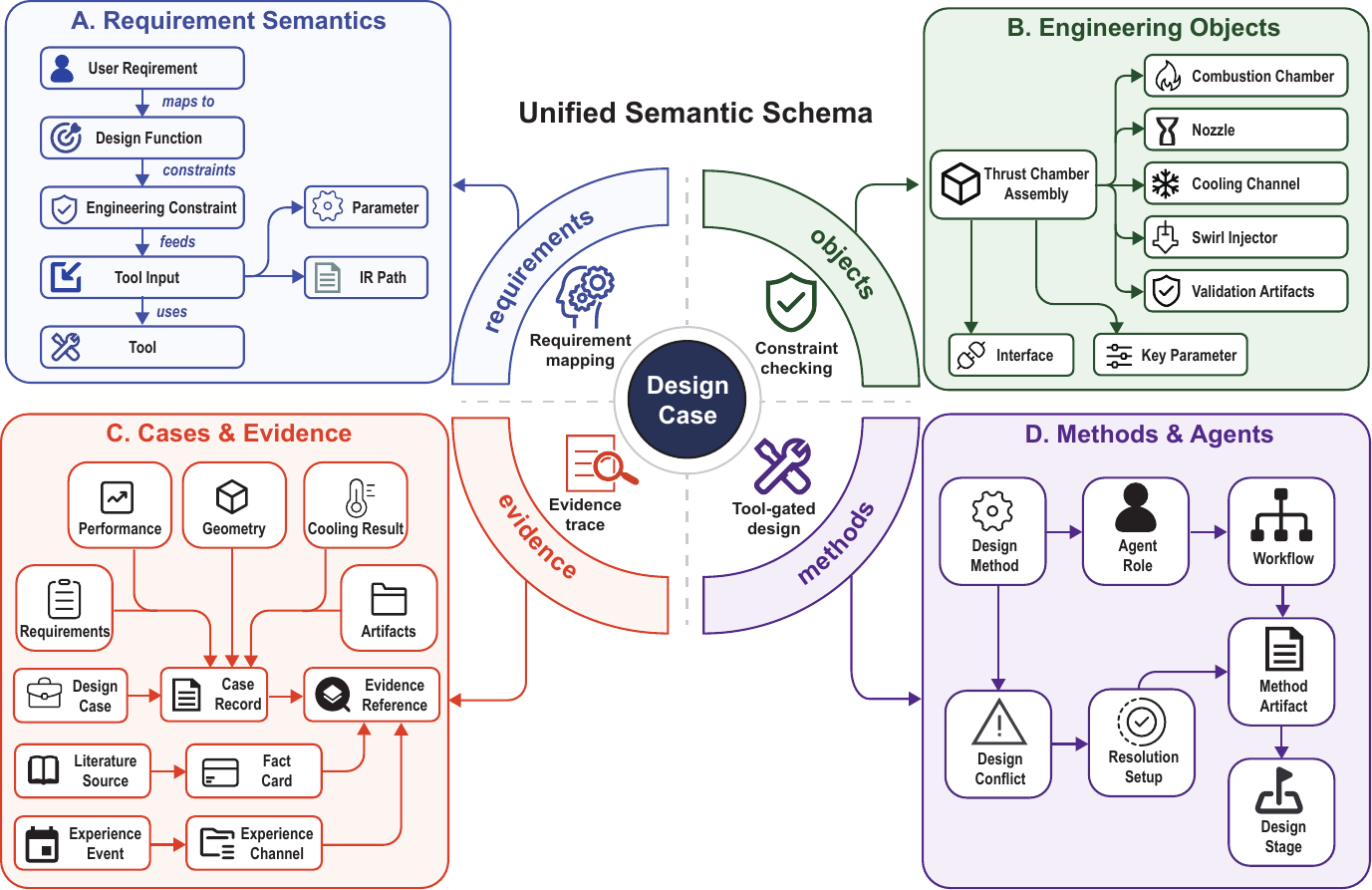}
\caption{KG semantic schema: (A) requirement semantics, (B) engineering objects, (C) cases and source-linked evidence, and (D) methods and agents. The four relation groups connect design intent to an accepted concept and numerical inputs.}
\label{fig:kg-semantic-schema}
\end{figure*}

The KG provides structured evidence for choosing design actions and binding their inputs. Huang et al.~\cite{huang2025kgarchitecture} similarly connect product, feature, and decision knowledge to architecture selection and simulation. In RocketAgent, Fig.~\ref{fig:kg-semantic-schema} organizes this connection around requirements and functions (region A), engineering objects and interfaces (B), source-linked cases and evidence (C), and methods and conflict-resolution steps (D). A query combines the current workflow stage, design-record identifier, and accepted requirements, including thrust, chamber pressure, propellants, nozzle configuration, and cooling need. These fields identify relevant knowledge before the Agent commits the next engineering action.

The response returns ranked prior designs, RPA examples, stage-compatible method cards, reviewed facts, and a bounded set of typed graph relations with source references. Similarity relations provide candidate precedents, while conflict relations expose competing objectives and connect them to possible mitigation or re-evaluation actions. This context helps the Agent select the next continuation without treating retrieved knowledge as a substitute for current engineering analysis. Ranking, stage selection, and relation extraction use the configured selectors and budgets reported in ~\ref{app:reproducibility}.

Before use, the Agent compares each item's operating context and provenance with the current task and records adoption or rejection. Deterministic checks enforce units, numerical ranges, and required inputs when the selected skill is invoked, while missing information remains an unresolved prerequisite. Human approval remains required for permissions and consequential changes; retrieval itself does not create authorization. Adopted and rejected knowledge, unresolved prerequisites, and resulting calculations are linked through provenance records in the design state.

An observed representative-case trace illustrates how this mechanism changes the next design action. For the accepted 500~N, 1.0~MPa LOX/RP-1 concept, the KG returned a historical cooling precedent together with a conflict between maximum specific impulse and cooling feasibility. The Agent treated the precedent as a bounded seed, rejected a current-case self-match as independent evidence, and checked the precedent against the current operating conditions. When the initial cooling search found no feasible candidate within its stated budget, the retrieved resolution guidance supported an authorized rich-side continuation. Fresh performance, contour, and cooling calculations then identified $r_{\mathrm{Ox/Fuel}}=1.09$ and eight hard-feasible candidates among 70 evaluations. Thus, the KG supplied context, constraints, and a recoverable continuation, while current numerical calculations determined the accepted design values; provenance and adoption/rejection decisions remain linked in the design record and are subject to review before reuse.

\subsection{Skill-Coordinated Embodiment Design}
\label{subsec:embodiment-design}

\begin{figure*}[!t]
\centering
\includegraphics[width=\textwidth]{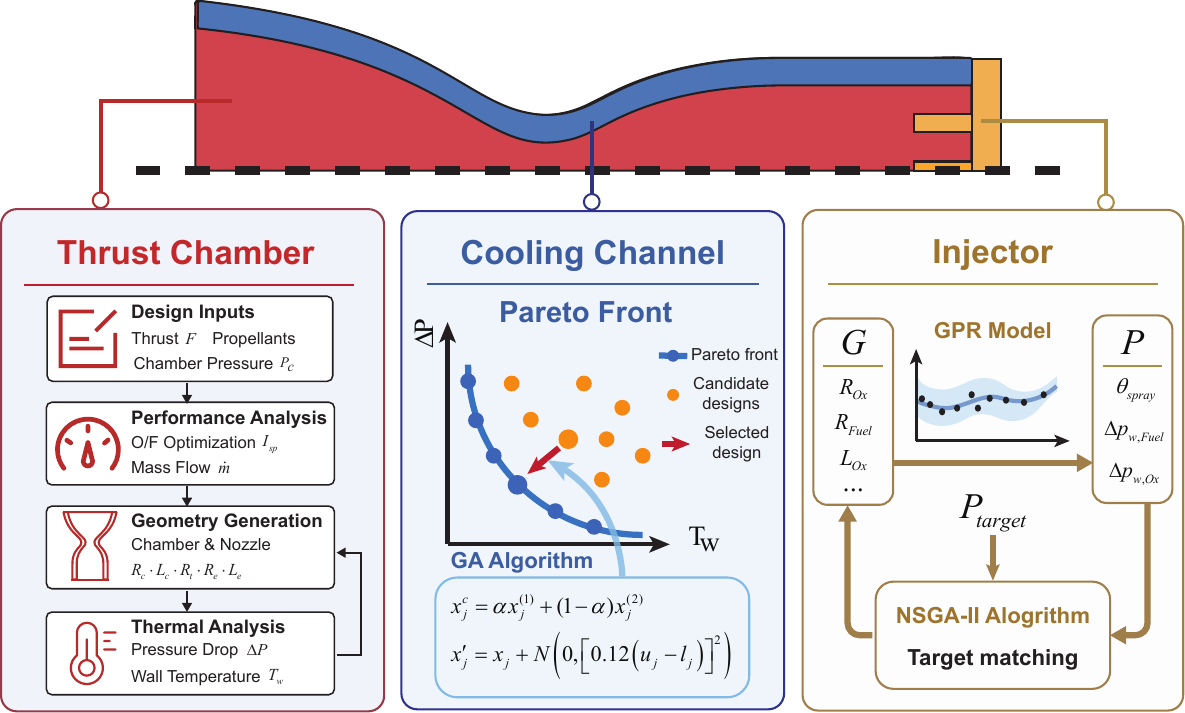}
\caption{Embodiment-design tasks mapped to the thrust-chamber regions. Left: performance and contour determination with thermal feedback. Center: GA crossover and mutation, a schematic cooling Pareto front, and candidate selection. Right: GPR prediction of water-reference injector responses and NSGA-II target matching. The cooling trade-off and GPR response curve are schematic.}
\label{fig:embodiment-design}
\end{figure*}

Fig.~\ref{fig:embodiment-design} links three embodiment-design task families to their physical regions in the thrust chamber. The left panel propagates thrust, propellant, and chamber-pressure inputs through mixture-ratio optimization, flow sizing, contour generation, and thermal analysis; the return arrow connects thermal feedback to geometry refinement. The central panel couples GA candidate generation to a schematic wall-temperature--pressure-loss projection of the cooling trade-off, with a selected candidate highlighted on the Pareto front. The right panel closes the injector search loop: GPR maps candidate geometry to water-reference spray-angle and pressure-drop predictions, and NSGA-II updates the geometry against prescribed response targets at fixed propellant flows. All three families share an operating state and must yield a compatible contour, hard-feasible cooling layout, and admissible injector before PicoGK realizes the assembled geometry.

\subsubsection{Thrust-Chamber Performance and Contour Determination}
\label{subsubsec:rocketcea-specific-impulse}
\label{subsubsec:rpa-geometry}

The accepted propellant pair, chamber pressure, ambient condition, and target thrust define the performance calculation. RocketCEA evaluates equilibrium specific impulse over the prescribed mixture-ratio interval and identifies the maximum-performance operating point \cite{gordon1994cea,taylor2026rocketcea}. When nozzle expansion is not prescribed, the exit pressure is matched to the ambient pressure. An optional, source-supported upper bound may reduce the specific impulse used for conservative flow sizing when comparable prior-design evidence is available. The raw calculation and the resulting design value are retained separately.

The design specific impulse $I_{\mathrm{sp}}$ determines total propellant flow, while the oxidizer-to-fuel ratio $r_{\mathrm{Ox/Fuel}}$ determines its allocation:
\begin{equation}
\dot{m}=\frac{F}{I_{\mathrm{sp}}g_0}, \qquad
\dot{m}_{\mathrm{Ox}}=\frac{r_{\mathrm{Ox/Fuel}}}{1+r_{\mathrm{Ox/Fuel}}}\dot{m}, \qquad
\dot{m}_{\mathrm{Fuel}}=\frac{1}{1+r_{\mathrm{Ox/Fuel}}}\dot{m}.
\label{eq:mass-flow-ledger}
\end{equation}
Here, $F$ is the target thrust, $g_0$ is standard gravitational acceleration, and the subscripts $\mathrm{Ox}$ and $\mathrm{Fuel}$ denote oxidizer and fuel. Eq.~\ref{eq:mass-flow-ledger} supplies the physical flow budget used by cooling and injector design. RPA-reported flow provides a cross-tool consistency check.

RPA sizes the chamber and nozzle from this operating point, the contraction ratio, characteristic length, and selected nozzle family \cite{ponomarenko2010rpa,rao1958nozzle,rezende2016parametric}. Supported contours include axisymmetric TOC, conical, and parabolic-bell profiles. The resulting chamber, throat, and exit dimensions are accompanied by an axially ordered contour shared by thermal analysis and geometry realization. Positive dimensions, increasing axial coordinates, and consistent geometric anchors are checked again after dimensional quantization. If cooling cannot accommodate the maximum-performance operating point, an engineer-authorized off-peak mixture ratio initiates a new performance, contour, and thermal evaluation. Thus, the performance optimum remains subject to subsystem feasibility.

\subsubsection{Regenerative-Cooling Design under Coupled Constraints}
\label{subsubsec:ga-cooling-screening}

Cooling design seeks thermally effective channel layouts without excessive pressure loss or inadequate wall strength. A bounded search varies channel count, inlet and outlet widths and depths, hot-wall thickness, channel angle, and coolant-flow factor. The genetic algorithm generates candidates through selection, crossover, and mutation; geometric checks eliminate unrealizable channel arrangements before thermal evaluation. In the GA inset of Fig.~\ref{fig:embodiment-design}, $x_j^{(1)}$ and $x_j^{(2)}$ are parental values of design variable $j$, and $\alpha$ is the arithmetic-crossover weight producing $x_j^c$. Gaussian mutation produces $x_j^{\prime}$ with perturbation standard deviation $0.12(u_j-l_j)$, where $l_j$ and $u_j$ are the variable bounds.

Selection prioritizes feasibility and then the configured scalar score. Separate nondominated postprocessing minimizes four normalized responses: maximum wall temperature, coolant outlet temperature, pressure loss, and structural risk. The first three are normalized by their respective limits; structural risk is $\mathrm{SF}_{\mathrm{req}}/\mathrm{SF}_{\min}$, where $\mathrm{SF}_{\mathrm{req}}$ is the required safety factor.

Each admitted candidate is evaluated using a single full-contour RPA channel-wall model, with fuel flowing from the nozzle exit toward the chamber head. RPA provides stationwise heat flux, wall temperatures, coolant states, and pressure loss. A low-order structural screen combines pressure loading, the through-wall thermal gradient, and local channel-span bending \cite{nasa1972fluidcooled,kim2014multidisciplinary}. At axial position $x$, its scalar stress is
\begin{equation}
\sigma_{\mathrm{scr}}(x)=\sigma_{\mathrm{vm,p}}(x)+\sigma_{\mathrm{th}}(x)+\sigma_{\mathrm{bend}}(x)+\sigma_{\mathrm{r}}(x),
\label{eq:low-order-stress}
\end{equation}
where $\sigma_{\mathrm{vm,p}}$ combines rib-ligament-corrected hoop pressure and closed-end axial pressure, $\sigma_{\mathrm{th}}=\alpha E|T_{\mathrm{wi}}-T_{\mathrm{wc}}|/[2(1-\nu)]$, $\sigma_{\mathrm{bend}}=|P_{\mathrm{cool}}-P_{\mathrm{g}}|w_{\mathrm{ch}}^2/(2t_{\mathrm{w}}^2)$, and $\sigma_{\mathrm{r}}$ is an optional mean thermal-restraint term. Here, $T_{\mathrm{wi}}$ and $T_{\mathrm{wc}}$ are the hot-side and coolant-side wall temperatures; $P_{\mathrm{g}}$ and $P_{\mathrm{cool}}$ are the gas and coolant pressures; $w_{\mathrm{ch}}$ and $t_{\mathrm{w}}$ are the channel span and hot-wall thickness; and $\alpha$, $E$, and $\nu$ are the thermal-expansion coefficient, Young's modulus, and Poisson's ratio. The minimum screening safety factor is
\begin{equation}
\mathrm{SF}_{\min}=\min_{x\in\Omega_{\mathrm{tc}}}\frac{\sigma_{\mathrm{allow}}(T_{\mathrm{w}}(x))}{\sigma_{\mathrm{scr}}(x)},
\label{eq:low-order-safety-factor}
\end{equation}
where $\Omega_{\mathrm{tc}}$ contains the sampled chamber/nozzle wall stations and $\sigma_{\mathrm{allow}}$ is the configured allowable stress. The screen supports preliminary chamber and nozzle wall-strength assessment. The representative design uses CuCrZr with the common material definition and component-specific limits listed in ~\ref{app:reproducibility}.

Downstream admission requires a valid channel-wall thermal solution and simultaneous satisfaction of wall-temperature, coolant-state, pressure-loss, manufacturing, and safety-factor limits. Fuel-specific coolant-temperature and coking screens are included where applicable. Nondominated ranking reveals trade-offs among evaluated candidates, but only a candidate passing the complete hard-constraint set can be selected for geometry realization. A favorable ranking alone cannot promote an infeasible design. Search bounds and numerical settings are retained in ~\ref{app:embodiment-settings}.

\subsubsection{Fixed-Flow Surrogate-Assisted Injector Screening}
\label{subsubsec:injector-screening}

Injector design allocates the accepted Ox and Fuel flows among a chosen number of coaxial swirl elements within the injector-face envelope and igniter-clearance requirement. Changing the element count changes the single-element flow, so each layout receives a separate geometry search. The externally trained GPR model uses ten geometric or derived features and two original-propellant single-element flow features. RocketAgent keeps these flow features fixed during optimization and does not retrain the model during the design execution.

The model-development VOF calculations use water in both liquid circuits. Their inlet flows are related to the original-propellant flows by
\begin{equation}
\dot{m}^{(\mathrm{e})}_{\mathrm{w},j}
=\frac{\dot{m}_{j}}{N_{\mathrm{e}}}\sqrt{\frac{\rho_{\mathrm{w}}}{\rho_j}},
\qquad j\in\{\mathrm{Ox},\mathrm{Fuel}\},
\label{eq:injector-water-equivalent-flow}
\end{equation}
where $N_{\mathrm{e}}$ is the element count and $\mathrm{w}$ denotes water. This mapping follows $\dot{m}=C_{\mathrm{d}}A\sqrt{2\rho\Delta p}$ and approximately preserves pressure drop for unchanged geometry and approximately equal discharge coefficients $C_{\mathrm{d}}$. Training features retain the original-propellant flows, whereas training responses correspond to the mapped water conditions. Consequently, RocketAgent supplies the original-propellant flows directly at inference without applying the conversion again. GPR predicts the water-reference spray-cone angle, Fuel- and Ox-side pressure drops, and their standard deviations.

The injector panel in Fig.~\ref{fig:embodiment-design} denotes the candidate geometry by $\boldsymbol{G}$ and the three predicted mean water-reference responses by $\boldsymbol{P}$. The target vector $\boldsymbol{P}_{\mathrm{target}}$ specifies the desired spray angle and Fuel- and Ox-side pressure drops. NSGA-II varies the geometry to minimize normalized, uncertainty-penalized deviations from the prescribed response targets \cite{deb2002nsgaii}. Numerical screening checks the feature domain, geometric constraints, a provisional mean spray-angle interval, and both mean water-reference pressure drops, which must exceed $0.20P_{\mathrm{c}}$. Dimensional and angular search variables are quantized and freshly evaluated. For the representative case, selection additionally requires both pressure-drop values $\mu-2\sigma$ to exceed the same threshold, followed by proximity to the spray-angle target and the minimum conservative pressure-drop margin. These selection values are not asserted to be calibrated confidence bounds \cite{kuleshov2018calibrated}. The full numerical settings and the distinction between mandatory checks, provisional engineering screening, optimization targets, and uncertainty reporting are given in ~\ref{app:embodiment-settings}. Each retained layout then proceeds to its own geometry realization and configuration-level verification. Held-out prediction errors and independent post-selection VOF comparisons are reported in Section~\ref{subsec:results-optimization}.

\textit{Common geometry-realization handoff.}
The three design families converge on a shared geometry-realization step. PicoGK combines the accepted chamber/nozzle contour, hard-feasible cooling layout, and admissible count-specific injector geometry into solid and fluid assemblies \cite{leap71picogk}. The RPA injector-face-to-exit coordinate convention is retained. Each layout produces its own STL assemblies and linked generation record. Coordinate consistency and artifact checks establish the handoff, while topology and mesh suitability require their own validation records. The resulting analysis-specific domains support the multiphysics verification described next. ~\ref{app:embodiment-settings} records the numerical settings and geometry-handoff checks.

\subsection{Sequential Verification and Candidate Screening}
\label{subsec:verification-refinement}
Verification is a linear sequence with three solver stages. First, DeepFlame VOF checks injector-scale spray and hydraulic response. Second, Fluent evaluates chamber combustion and flow distribution, including Ox-channel total-flow consistency and the resolved Ox branch distribution. Third, only candidates that pass all preceding checks enter Abaqus for thermal--structural strength assessment. Failure of any verification check rejects the current candidate immediately; no downstream solver is invoked for that candidate. Each stage retains its evidence and disposition in the shared design state. Case-specific meshes, boundary conditions, material cards, solver controls, and numerical outputs are reported in \ref{app:reproducibility}, ~\ref{app:simulation-settings} and in Section~\ref{sec:evaluation-results}.

\subsubsection{Injector Spray and Hydraulic Response}
\label{subsubsec:terminal-evidence-roles}
DeepFlame is selected for injector-scale evidence because its OpenFOAM-derived finite-volume infrastructure supports transient volume-of-fluid (VOF) interface transport and the resolution of liquid-sheet development and primary breakup \cite{mao2023deepflame,deshpande2012interfoam,hirt1981vof,natarajan2024biswirl}. The resulting phase-fraction and velocity fields complement scalar spray-angle and pressure-drop responses, while the solver's fluxes provide a direct conservation check. Turbulence and pressure--velocity coupling are configured through the declared backend interface. Passing this VOF stage is a prerequisite for the Fluent assessment.For each endpoint response $j$, the VOF consistency deviation is defined as $\delta^{\mathrm{VOF}}_j=(\mu^{\mathrm{GPR}}_j-y^{\mathrm{VOF}}_j)/y^{\mathrm{VOF}}_j$, where $y^{\mathrm{VOF}}_j$ is the independent VOF window mean and therefore the numerical reference for this comparison. The VOF gate requires $|\delta^{\mathrm{VOF}}_j|\leq5\%$ for spray angle and for the Fuel- and Ox-side pressure drops. This gate compares the surrogate prediction with the independently calculated VOF response.

\subsubsection{Fluent Combustion Assessment}
After VOF acceptance, a chamber-scale reacting-flow solver is required because injector-level interface dynamics do not determine the coupled temperature and pressure fields in the chamber and nozzle. Fluent is used for this role because it supports three-dimensional finite-volume reacting-flow calculations, turbulence and spray closures, combustion-chemistry models, and field or integral post-processing on the realized geometry. The resulting temperature, pressure, and integral responses provide the combustion-side evidence used in the preliminary-design screen. For each integral response $y_j$, the relative deviation is defined as $\delta_j=(y_j-y_j^{\mathrm{ref}})/y_j^{\mathrm{ref}}$, where $y_j^{\mathrm{ref}}$ is the corresponding accepted-design or RocketCEA reference; the combustion gate requires $|\delta_j|<10\%$ for chamber pressure, thrust, specific impulse, and characteristic velocity. This threshold is an engineering consistency criterion for preliminary configuration screening. A candidate that fails the combustion check is discarded; passing it permits evaluation of the Fluent flow-distribution evidence. Its configuration, chemistry, and discretization details are retained in ~\ref{app:simulation-settings}.

\subsubsection{Fluent Flow-Distribution Screening}
Flow-distribution screening completes the Fluent stage before any Abaqus calculation. The Ox-channel flow check compares the sum of integrated Ox branch outlet flows with the prescribed Ox inlet flow, while the individual branch values determine the within-layout distribution. Both checks must pass for a candidate to advance to structural assessment. For a resolved Oxidizer feed, let $\dot{m}_{\mathrm{Ox},i}$ be the integrated outlet flow of element $i$, and let $\overline{\dot{m}}_{\mathrm{Ox}}$ be the layout mean. The coefficient of variation and maximum relative deviation are
\begin{equation}
\begin{array}{l}
\displaystyle \mathrm{CV}_{\mathrm{Ox}}=\frac{\sqrt{N_{\mathrm{e}}^{-1}\sum_{i=1}^{N_{\mathrm{e}}}(\dot{m}_{\mathrm{Ox},i}-\overline{\dot{m}}_{\mathrm{Ox}})^2}}{\overline{\dot{m}}_{\mathrm{Ox}}}\times100\%, \\[6pt]
\displaystyle \delta_{\mathrm{Ox},\max}=\max_i\frac{|\dot{m}_{\mathrm{Ox},i}-\overline{\dot{m}}_{\mathrm{Ox}}|}{\overline{\dot{m}}_{\mathrm{Ox}}}\times100\%.
\end{array}
\label{eq:injector-flow-uniformity}
\end{equation}
The flow-distribution gate requires both $\mathrm{CV}_{\mathrm{Ox}}\leq5\%$ and $\delta_{\mathrm{Ox},\max}\leq5\%$. A layout fails if either measure exceeds its limit. Passing the Ox total-flow check alone does not establish acceptable branch distribution. A rejected layout is discarded before Abaqus. The flow-admission definitions and the configuration-specific values are retained in ~\ref{app:reproducibility} and Section~\ref{subsec:results-multiphysics}.

\subsubsection{Coupled Thermal--Structural Strength}
Abaqus provides the final verification stage for candidates that have passed both VOF and the Fluent combustion and flow-distribution checks. It is selected because it provides finite-element thermal analysis, pressure loading, support conditions, and transfer of temperatures into a mechanical step within one solver environment. The injector top plate is treated as the focal structural region because it carries the coupled loads around the injector and igniter openings; this makes a declared stress path more interpretable than an unqualified maximum over all constrained edges. A structural failure discards the candidate; passing the structural check completes the declared preliminary top-plate structural assessment and permits final selection. The assessment records temperature, displacement, stress, and reaction-force evidence, with the material definition and admission criteria kept in ~\ref{app:reproducibility}.

\subsubsection{Candidate Rejection and Design-State Governance}
\label{subsubsec:failure-diagnosis-recovery}
\label{subsubsec:human-approval-knowledge-evolution}
A failed verification check returns its violated criterion and supporting evidence to the Agent, which rejects the candidate and cancels its remaining verification tasks. The failed outcome remains available for engineering review. Design revision is a separate operation governed by approval: when an upstream design-state change is authorized, dependency-scoped invalidation marks affected results stale and determines which prerequisites must be re-established. This revision mechanism does not override the linear pass-or-reject logic of candidate verification.

The same Agent also drives the simulation stages through the declared skills and tool adapters: it selects the eligible solver task, binds the accepted Design IR to solver inputs, launches or coordinates the backend run, inspects the returned evidence, and advances to the next verification stage only on a pass, and discards the candidate on failure. The Agent-mediated CFD procedure follows the method developed in our team's previous FlamePilot work, which uses an LLM-based research assistant to configure, execute, post-process, and iteratively refine DeepFlame/OpenFOAM simulations \cite{xiao2026flamepilot}. RocketAgent adopts this procedure directly and extends it across the ordered injector VOF, Fluent combustion and flow-distribution, and Abaqus structural stages within a shared design state. Human approval governs requirements, consequential model or safety-limit changes, expensive simulation commitments, and final selection. Verified, failed, and unresolved outcomes are retained as separate evidence states, while promotion of a result into reusable engineering knowledge additionally requires provenance, applicability, and expert review. This governance layer connects solver outputs to design continuation. Execution traces and numerical settings are provided in ~\ref{app:agent-execution} and \ref{app:simulation-settings}; the representative outcomes and revision-governance checks are reported in Section~\ref{sec:evaluation-results}.

\section{Performance Evaluation and Results}
\label{sec:evaluation-results}

\begin{figure*}[!t]
\centering
\includegraphics[width=\textwidth]{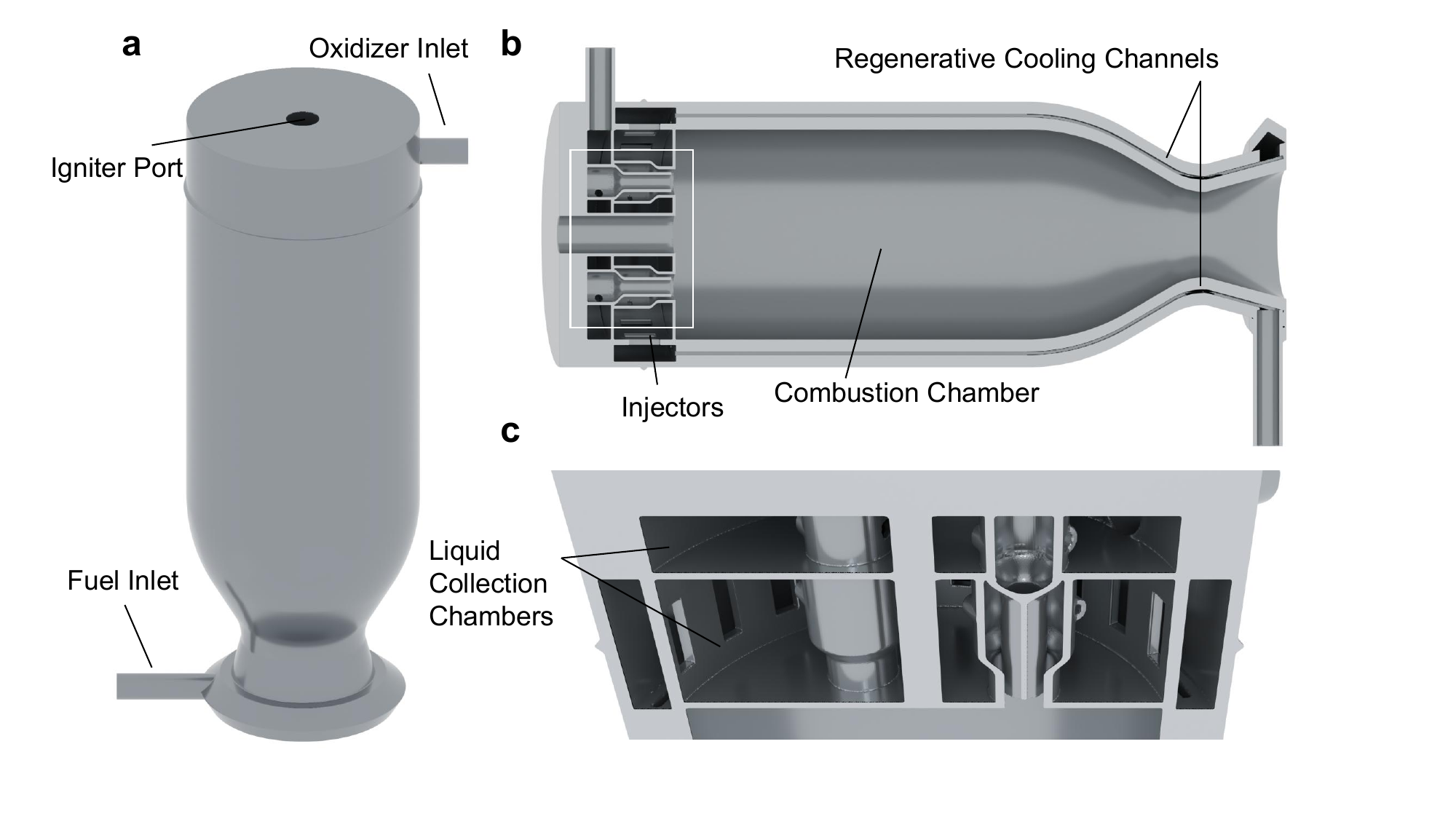}
\caption{Selected four-element thrust-chamber geometry: (a) exterior view showing the igniter port and separate oxidizer and fuel inlets; (b) longitudinal section showing the injectors, combustion chamber, nozzle, and regenerative-cooling channels; (c) enlarged injector-head cutaway showing the liquid collection chambers and internal injector passages.}
\label{fig:representative-geometry}
\end{figure*}

\textit{Result at a glance.}
RocketAgent transformed engineer-confirmed requirements into the selected, multiphysics-evaluated four-element LOX/RP-1 thrust chamber shown in Fig.~\ref{fig:representative-geometry}. The resulting artifact combines a cooling-feasible operating point, an embodied injector and nozzle, solver-ready solid and fluid geometry, and evidence from the ordered VOF, Fluent, and Abaqus stages, all connected through one shared design state. This section follows the execution order: it first establishes the accepted conceptual basis, then reports embodiment calculations and candidate realization, and finally presents the sequential verification and selection results. The execution record also captures the coordination work between calculations, including an authorized continuation after cooling-search failure, correction of an input-binding inconsistency, and retention of unresolved geometry work in the plan.

The delivered assembly connects the external interfaces to the internal geometry developed across the workflow. Panel (a) locates the central igniter port, head-end Ox inlet, and nozzle-end Fuel inlet. The section in (b) exposes the injector head, chamber/nozzle contour, and regenerative-cooling passages, while the cutaway in (c) reveals the collection chambers and internal injector passages. The following results explain how this integrated design was obtained and why it was selected.

\begin{figure*}[!t]
\centering
\includegraphics[width=\textwidth,height=0.75\textheight,keepaspectratio]{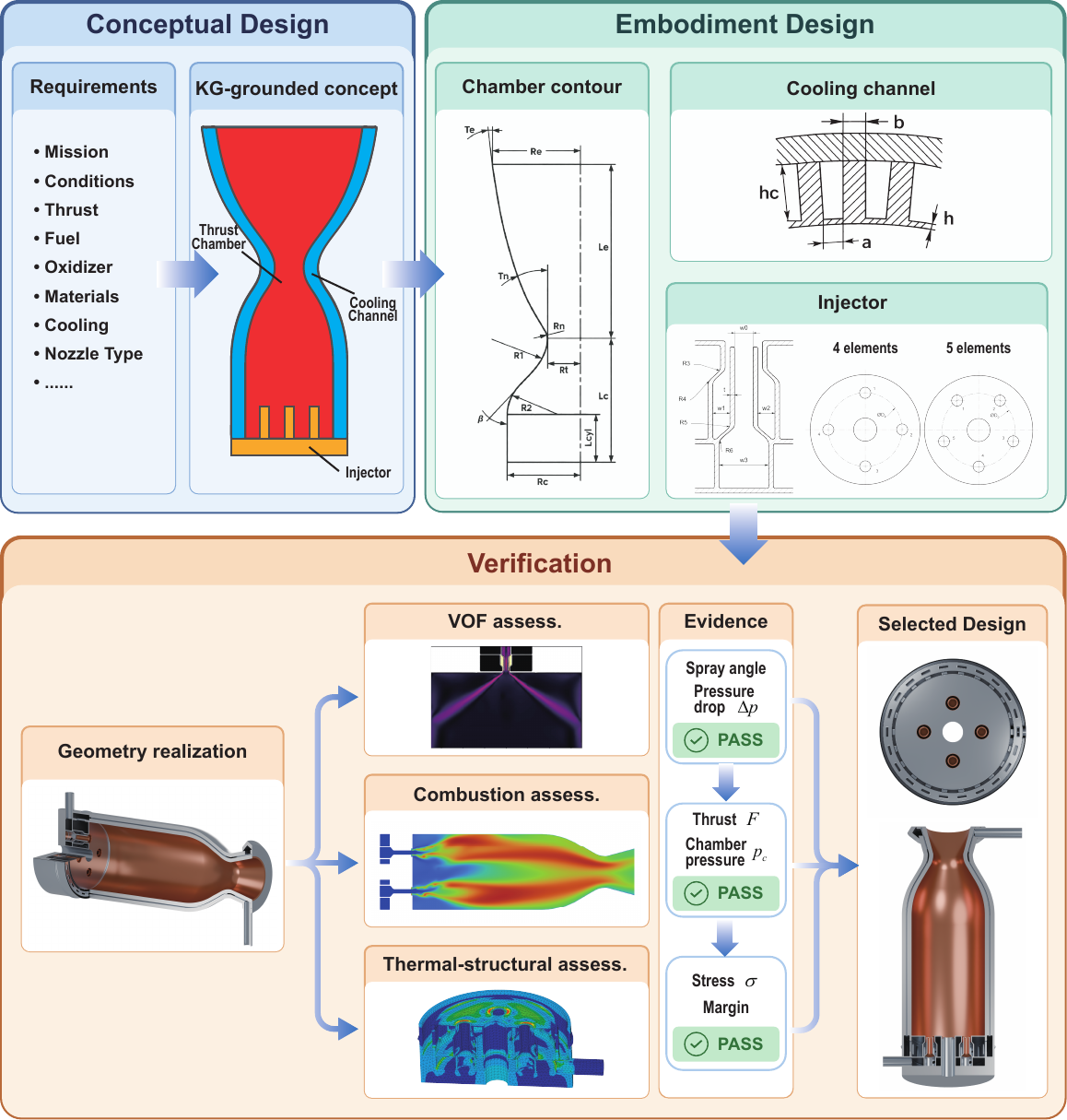}
\caption{Representative 500~N LOX/RP-1 design sequence. The upper row shows conceptual and embodiment design; the lower band groups the realized geometry and verification evidence by physical task.}
\label{fig:design-revision-transitions}
\end{figure*}

\textit{Workflow overview.}
Fig.~\ref{fig:design-revision-transitions} provides the roadmap for these results. Its upper row follows the transition from the accepted conceptual basis to the embodiment branches, while its lower band groups the realized geometry and verification evidence by physical task. The next subsections follow the same route: conceptual requirements and KG grounding, performance and embodiment calculations, and then verification through injector VOF, Fluent combustion and flow-distribution checks, and Abaqus thermal--structural checking. A candidate is discarded at the first failed verification stage; accordingly, the five-element layout stopped at the Fluent flow-distribution check, while the four-element layout completed the structural check and became the selected configuration.

\subsection{Conceptual Design: Requirements and Design Basis}
\label{subsec:evaluation-protocol}
The representative task requires a 500~N thrust chamber at sea level and $P_{\mathrm{c}}=1.0$~MPa, using LOX/RP-1, a $15^{\circ}$ conical nozzle, regenerative cooling, and CuCrZr. These accepted inputs instantiate the problem formulation in Section~\ref{subsec:problem-state}; mixture ratio, cooling geometry, and injector geometry remain design decisions. The knowledge-use procedure in Section~\ref{subsec:kg-conceptual-design} returned historical case and the maximum-specific-impulse/cooling conflict as relevant context. The Agent recorded the prior's limited applicability and rejected the current-case self-match as independent precedent. The KG trace described above shows how the historical cooling geometry became an actual scan input while current calculations determined the design values. These accepted inputs define the conceptual baseline passed to embodiment. The next subsection examines how performance, cooling, injector, and geometry calculations transform this baseline into realized candidates.

\subsection{Embodiment Design: From Accepted State to Realized Hardware}
\label{subsec:results-optimization}
With the conceptual state fixed, embodiment begins by asking whether the performance point can be realized thermally and hydraulically; only then do injector sizing and geometry generation proceed. RocketAgent carries the accepted state into a coupled cooling and injector design. The same Ox and Fuel totals are passed to both injector layouts, while the eventual cooling layout supplies the wall and flow context for geometry realization.

\subsubsection{Feasibility-Driven Operating-Point Revision}
\label{subsubsec:results-end-to-end}
The first embodiment output was a cooling-feasible operating state. The question was whether the maximum-performance point could support a feasible cooling design. Following the accepted concept and the knowledge-use procedure in Section~\ref{subsec:kg-conceptual-design}, the design record supplied the propellants, operating conditions, nozzle family, and cooling requirement to RocketCEA and RPA. The Agent's recorded plan retained geometry generation as pending until the subsystem checks were resolved. Conditional human authorization allowed a richer mixture at fixed chamber pressure if the bounded cooling search found no hard-feasible candidate (~\ref{app:agent-execution}).

At the maximum-$I_{\mathrm{sp}}$ operating point, none of 74 recorded cooling-candidate evaluations satisfied the complete thermal-hydraulic, manufacturing, and low-order structural criteria within the configured search space and evaluation budget. This establishes that the bounded search found no feasible candidate, rather than proving that no physically feasible cooling design exists at that operating point. The search outcome activated the authorized return from cooling assessment to performance and contour determination. The Agent invoked the rich-side search using the prior-case seed and bounded continuation described above, and recalculated the dependent performance, flow, geometry, and cooling responses. At $r_{\mathrm{Ox/Fuel}}=1.09$, eight of 70 recorded candidate evaluations were hard-feasible.

The revised operating point established the common flow budget for the remainder of embodiment and verification. Eq.~\ref{eq:mass-flow-ledger} gives $\dot{m}=0.282944$~kg/s, comprising 0.147564~kg/s Ox and 0.135380~kg/s Fuel. Thus, the mixture-ratio decision changed the inputs to cooling and injector design together, implementing the cross-stage dependency described in Section~\ref{subsec:overall-framework}.

\begin{figure*}[!t]
\centering
\includegraphics[width=0.55\textwidth]{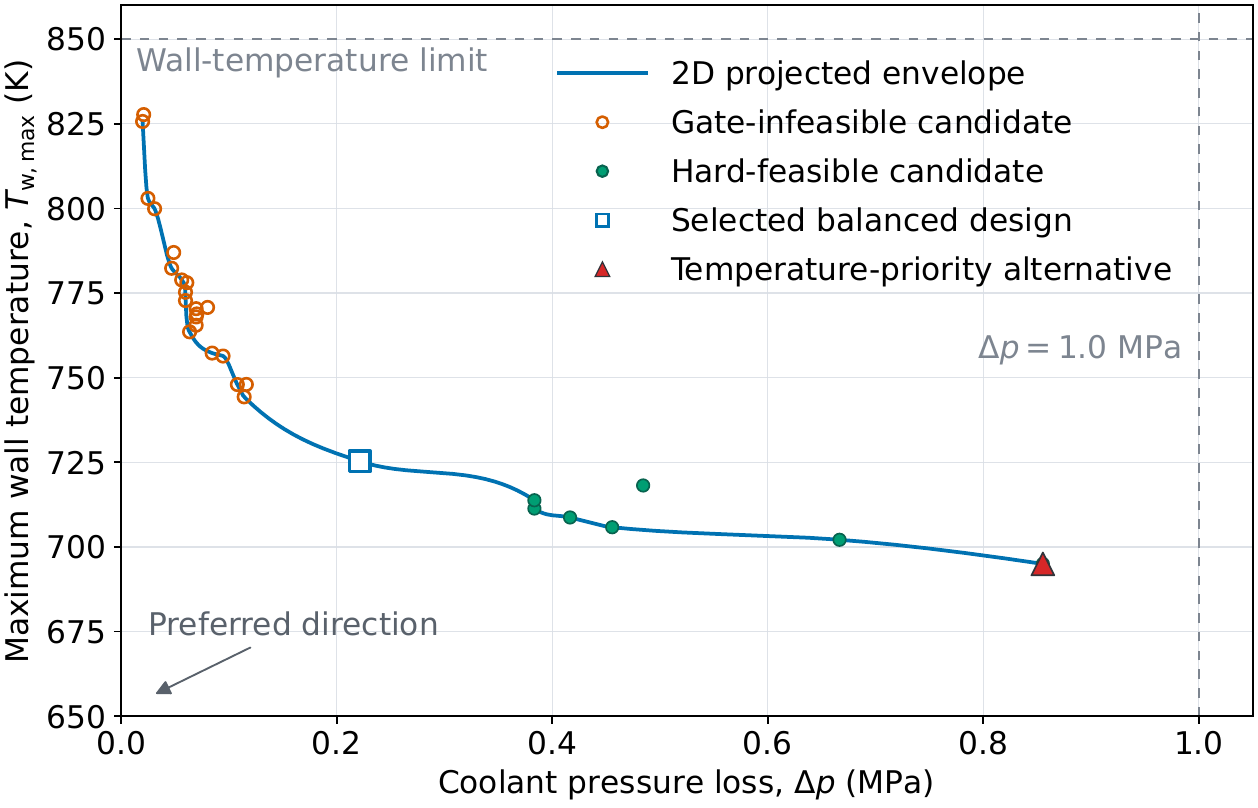}
\caption{Cooling-channel trade-off after the governed rich-side redesign. The axes show coolant pressure loss $\Delta p$ and maximum wall temperature $T_{\mathrm{w,max}}$. Hollow markers fail at least one hard gate, whereas green markers are fully feasible. The blue square marks the selected balanced design; the red triangle marks the temperature-priority alternative.}
\label{fig:cooling-pareto}
\end{figure*}

Fig.~\ref{fig:cooling-pareto} shows the trade-off remaining after this operating-point revision. Lower maximum wall temperature generally comes at greater coolant pressure loss along the projected envelope. The hollow orange points lie within the two displayed limits but fail another hard constraint, so these axes alone cannot determine admission. Among the hard-feasible alternatives, the blue square marks the selected balanced design and the red triangle a temperature-priority alternative. The selected balanced design uses a pressure loss of 0.221~MPa, a maximum wall temperature of 725.3~K, and a minimum low-order safety factor of 1.80 under the common CuCrZr definition. This balanced point was preferred because it retains substantial thermal and structural margin without the large pressure-loss penalty of the temperature-priority alternative, and it fixed the cooling state inherited by the injector and geometry branches.

With the operating point and cooling state fixed, the remaining embodiment task is to generate injector alternatives and realize their common geometry interfaces.

\subsubsection{Surrogate-Assisted Injector Embodiment}

\begin{figure*}[!t]
\centering
\includegraphics[width=0.94\textwidth]{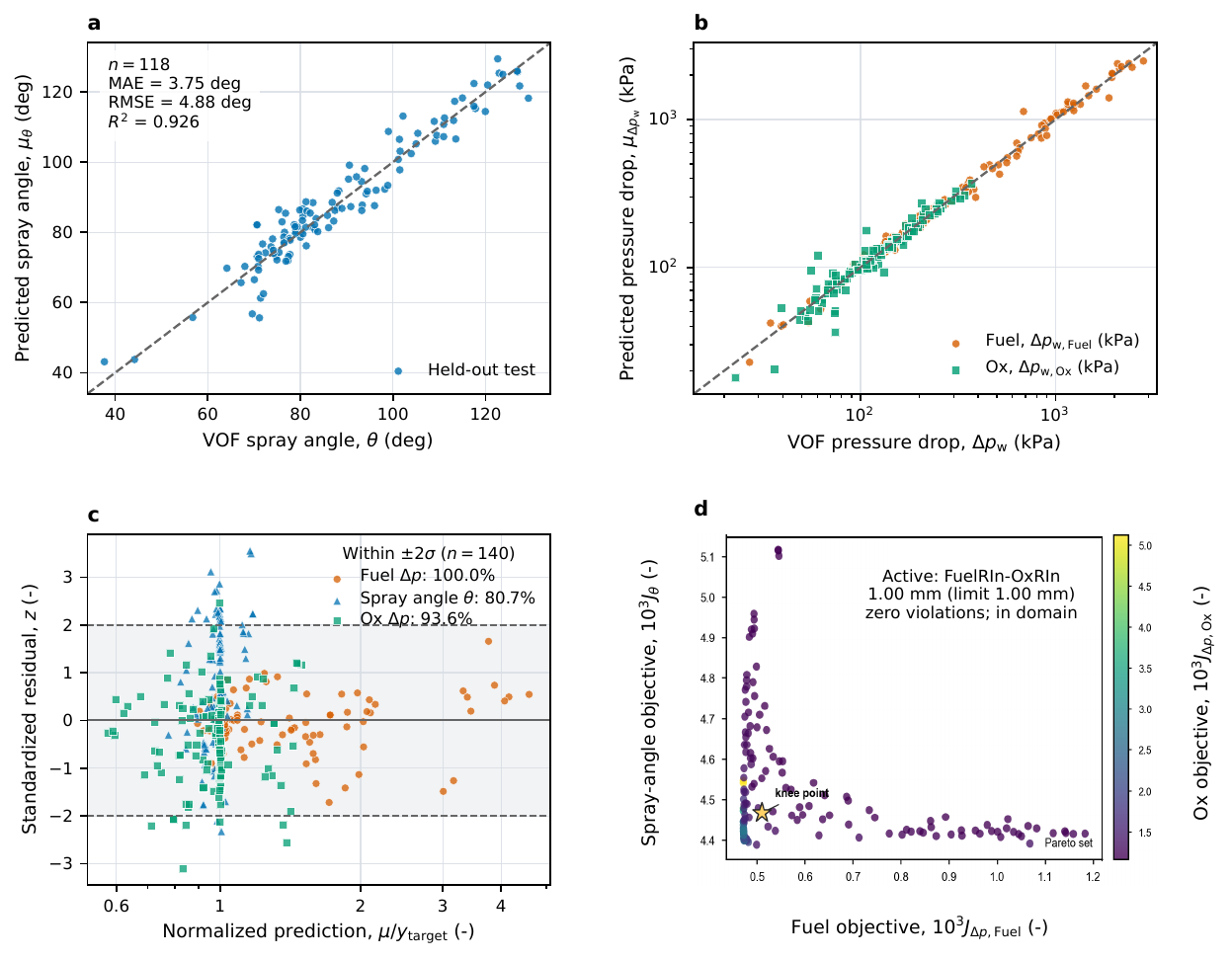}
\caption{Water-reference injector-module surrogate and search evidence. (a) Held-out VOF--surrogate parity for spray-cone angle ($n=118$). (b) Held-out parity for Fuel- and Ox-side pressure drops. (c) Standardized residuals $z=(y-\mu)/\sigma$ for 140 independent, search-conditioned post-selection VOF evaluations; dashed lines at $z=\pm2$ mark nominal two-standard-deviation limits, whose empirical coverage is reported in Table~\ref{tab:gpr-calibration}. (d) NSGA-II nondominated objective set, colored by the Ox-pressure-drop objective, with the selected module-validation knee point highlighted.}
\label{fig:injector-surrogate-evidence}
\end{figure*}

Fig.~\ref{fig:injector-surrogate-evidence} summarizes the prediction evidence supporting the count-specific searches. The upper panels compare held-out point predictions with VOF responses; the lower panels examine post-selection residuals and an optimization trade-off. These evaluations define the injector embodiment search and its uncertainty context; the independent endpoint VOF verification of the realized configuration is reported in the verification subsection.

Injector-module validation and the representative-design searches used the same trained GPR surrogate, with 549 training samples, 118 validation samples, and 118 non-overlapping held-out VOF samples, all based on water-substitute calculations. The common model configuration is detailed in ~\ref{app:embodiment-settings}. The reported NRMSE values use the training-response ranges defined in Eq.~\ref{eq:gpr-nrmse}. In Fig.~\ref{fig:injector-surrogate-evidence}(a), distance from the diagonal measures spray-angle prediction error; the complete held-out set achieved $R^2=0.926$, $\mathrm{MAE}=3.746^{\circ}$, $\mathrm{RMSE}=4.881^{\circ}$, and $\mathrm{NRMSE}=4.314\%$. Panel (b) compares the two pressure drops on logarithmic axes, accommodating their different response ranges. Fuel-side pressure drop achieved $\mathrm{MAE}=42.043$~kPa, $\mathrm{RMSE}=89.123$~kPa, and $\mathrm{NRMSE}=3.247\%$; Ox-side pressure drop achieved $\mathrm{MAE}=10.133$~kPa, $\mathrm{RMSE}=16.234$~kPa, and $\mathrm{NRMSE}=4.106\%$. These statistics retain all 118 held-out samples and quantify the point-prediction evidence behind the upper panels.

Seven target-conditioned NSGA-II searches used a population of 160 for 100 generations and recorded 16,160 surrogate evaluations per target, or 113,120 in total. They produced 1,079 feasible first-front candidates and 140 exported representatives. After the candidate list and search decisions were frozen, these 140 representatives were evaluated in separate VOF runs. The resulting labels were not used for GPR fitting, calibration, ranking, or selection; they therefore form an independent post-selection verification set for this search. The surrogate-to-VOF spray-angle MAE/RMSE was $5.895^{\circ}/7.476^{\circ}$, the Fuel-pressure-drop MAE/RMSE was 17.905/24.485~kPa, and the Ox-pressure-drop MAE/RMSE was 8.506/10.691~kPa.

The 140 labels are independent of model fitting and search decisions, but they are search-conditioned because the evaluated geometries were exported by the surrogate-assisted procedure rather than sampled randomly from the full design domain. Fig.~\ref{fig:injector-surrogate-evidence}(c) uses the predictions recorded before those VOF runs to separate target matching from uncertainty: its horizontal coordinate is prediction divided by target, whereas its vertical coordinate is the standardized VOF residual. A prediction near the target therefore need not have a small residual. For response $j$, $z_{ij}=(y_{ij}-\mu_{ij})/\sigma_{ij}$ and empirical coverage is $C_j(k)=n^{-1}\sum_i\mathbf{1}(|z_{ij}|\leq k)$, with $n=140$. Here, $\sigma_{ij}$ is the GPR predictive standard deviation, including its fitted noise-kernel contribution. Table~\ref{tab:gpr-calibration} reports the resulting search-conditioned uncertainty statistics, and Fig.~\ref{fig:gpr-coverage} provides the calibration curves.

\begin{table*}[!t]
\centering
\footnotesize
\setlength{\tabcolsep}{6pt}
\renewcommand{\arraystretch}{1.08}
\caption{GPR uncertainty statistics for 140 independent post-selection VOF evaluations.}
\label{tab:gpr-calibration}
\begin{tabular}{@{}lrrrrr@{}}
\hline
\textbf{Response} & $C(1)$ (\%) & $C(2)$ (\%) & \textbf{Mean $z$} & \textbf{SD $z$} & \textbf{NLPD} \\
\hline
Spray-cone angle & 50.71 & 80.71 & 0.750 & 1.175 & 3.565 \\
$\Delta p_{\mathrm{w},\mathrm{Fuel}}$ & 92.86 & 100.00 & $-0.007$ & 0.540 & 11.762 \\
$\Delta p_{\mathrm{w},\mathrm{Ox}}$ & 64.29 & 93.57 & $-0.430$ & 0.945 & 10.680 \\
\hline
\end{tabular}
\end{table*}

The spray-angle markers in panel (c) extend beyond the nominal $\pm2\sigma$ band more frequently than the pressure-drop markers, consistent with their 80.71\% coverage and positive residual bias. Fuel-pressure intervals are conservative, while Ox-pressure intervals are closer to nominal coverage. This response-dependent coverage cautions against interpreting nominal GPR intervals as calibrated confidence bounds. The empirical coverage describes the evaluated search-conditioned population and does not guarantee coverage for an individual design.

Fig.~\ref{fig:injector-surrogate-evidence}(d) shows how a module-validation candidate was selected: the axes express Fuel-pressure and spray-angle target errors, color encodes the Ox-pressure objective, and the star marks the knee compromise. Against targets of 320~kPa, $80^{\circ}$, and 120~kPa, its independent VOF evaluation returned 317.06~kPa, $82.48^{\circ}$, and 127.11~kPa, respectively. The target-normalized RMSE of its three surrogate-to-VOF deviations was 3.93\%, and the maximum relative discrepancy between measured and prescribed water-circuit flow at $t=0.1$~s was 1.212~ppm. This module-level selection is separate from the representative-design $P_{\mathrm{c}}=1.0$~MPa dual-pressure-drop admission test.

\subsubsection{Count-Specific Design Alternatives}
The representative-design searches used the same trained GPR surrogate as the module validation above to generate separate count-specific embodiment candidates. Each candidate was quantized and freshly screened against the representative-case criteria stated in ~\ref{app:embodiment-settings} before geometry realization.
For the representative chamber, both sets of prescribed single-element propellant flows lay within the numerical flow-feature bounds of the common surrogate. Fixed-flow searches produced nondominated alternatives, which were quantized and freshly evaluated against the complete injector admission criteria. At the four-element allocation, the retained geometry had a predicted water-reference spray angle of $87.621^{\circ}$ and Fuel- and Ox-side mean pressure drops of 311.267 and 253.004~kPa, respectively. It satisfied the numerical feature-domain, provisional spray-angle, and dual-pressure-drop screens. Predictive standard deviations were $5.523^{\circ}$, 44.963~kPa, and 11.646~kPa for these three responses. The corresponding pressure-drop $\mu-2\sigma$ values were 221.34 and 229.71~kPa, both above the 200-kPa screening threshold.

The five-element alternative was optimized at its own single-element flows rather than reusing the four-element geometry. Its predicted water-reference spray angle was $85.993^{\circ}$, and its Fuel- and Ox-side mean pressure drops were 378.427 and 225.462~kPa. Predictive standard deviations were $5.686^{\circ}$, 44.035~kPa, and 11.031~kPa, respectively. The corresponding $\mu-2\sigma$ selection values were 290.36 and 203.40~kPa. Both count-specific designs passed the mean pressure-drop screen, the provisional spray-angle window, and the additional conservative pressure-drop condition. They were retained for geometry realization and independent verification, which determined the final configuration selection.

Independent endpoint VOF calculations provide the numerical reference for these count-specific candidates. Using the VOF window mean as denominator, the six absolute GPR--VOF relative deviations are 3.48\%, 0.65\%, and 2.64\% for the four-element allocation and 2.02\%, 0.12\%, and 1.33\% for the five-element allocation, listed in the order of spray angle, Fuel pressure drop, and Ox pressure drop. The table reports the corresponding differences $y_{\mathrm{VOF}}-\mu$, VOF-normalized relative deviations, and standardized residuals $z=(y_{\mathrm{VOF}}-\mu)/\sigma$; all values are computed before rounding. All six deviations satisfy the declared $5\%$ VOF gate, and the corresponding standardized residuals remain below 0.6. These results establish endpoint consistency with the independent VOF responses before either layout enters the Fluent stage.

\subsection{Verification: Sequential Screening and Final Acceptance}
\label{subsec:results-multiphysics}
The two numerically screened injector designs mark the end of embodiment search: both have passed the representative-case selection criteria, but neither is the final selected layout yet. They were combined with the accepted contour and cooling layout to produce separate PicoGK solid and fluid assemblies. This handoff implements the joint subsystem prerequisite in Section~\ref{subsec:embodiment-design}. Verification then proceeds in the fixed order VOF, Fluent combustion and flow distribution, and Abaqus strength checking. A candidate is discarded as soon as any check fails. The five-element layout was rejected for its flow distribution during the Fluent stage; only the four-element layout proceeded to Abaqus and completed verification. The following subsections follow this execution order.

\subsubsection{Injector VOF Endpoint Verification}
The endpoint checks compare both quantized injector candidates reported above with separate single-injector DeepFlame VOF calculations for the four- and five-element flow allocations. Table~\ref{tab:vof-endpoint} uses the same GPR means and predictive standard deviations as the candidate-admission results, alongside the VOF window means. For the five-element allocation, the VOF mean spray angle and Fuel- and Ox-side pressure drops are $87.767^{\circ}$, 378.882~kPa, and 228.493~kPa, respectively. These are water-reference injector responses; the chamber-scale reacting-flow results are assessed separately below. All six absolute standardized residuals are below 0.6, so both candidates' VOF responses lie within one predictive standard deviation of the corresponding GPR estimates.

The primary comparisons each use ten equally spaced samples from 0.091 to 0.100~s. For the four-element allocation, three samples carried a quality flag; excluding them reduced the mean spray-cone angle from $84.674^{\circ}$ to $83.396^{\circ}$. For the five-element allocation, two samples carried the same flag; the eight unflagged samples give $87.368^{\circ}$, compared with $87.767^{\circ}$ for the full window. These differences expose the effect of frame selection on the extracted spray-angle statistic without changing the retained full-window result. Case identifiers, sampling, and flag handling are detailed in ~\ref{app:simulation-settings}.

\begin{figure*}[!t]
\centering
\includegraphics[width=0.94\textwidth]{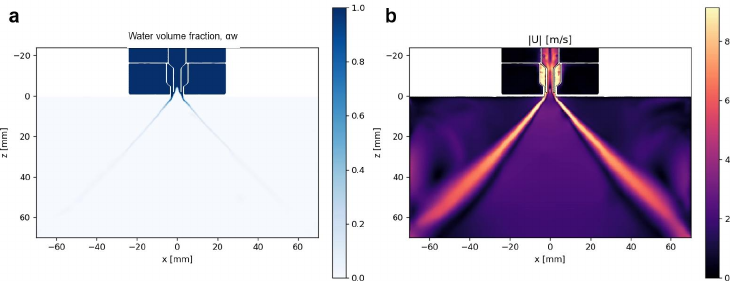}
\caption{Single-injector water-reference VOF fields for the four-element flow allocation. (a) Water volume fraction $\alpha_{\mathrm{water}}$ and (b) velocity magnitude $|\mathbf{U}|$ in m/s on the centerline $x$--$z$ plane at $t=0.1$~s.}
\label{fig:injector-vof-field}
\end{figure*}

Fig.~\ref{fig:injector-vof-field} retains the spatial evidence for the realized single injector at the four-element flow allocation. Panel (a) shows the centerline water-volume-fraction field and panel (b) the velocity magnitude at $t=0.1$~s, with the injector passages above the discharge plane and the spray region below. The scalar comparisons for both allocations are reported separately in Table~\ref{tab:vof-endpoint}. This injector-scale VOF evidence is the first verification layer before the chamber-scale Fluent and Abaqus assessments.

\begin{table*}[!t]
\centering
\footnotesize
\setlength{\tabcolsep}{5pt}
\renewcommand{\arraystretch}{1.08}
\caption{Independent GPR--DeepFlame VOF responses for the four- and five-element injector candidates.}
\label{tab:vof-endpoint}
\begin{tabular}{@{}lccccc@{}}
\hline
\textbf{Response} & \textbf{GPR $\mu\pm\sigma$} & \textbf{VOF mean} & \textbf{VOF--GPR} & \textbf{Relative deviation} & \textbf{$z$} \\
\hline
\multicolumn{6}{@{}l}{\textit{Four-element allocation}} \\
Spray angle & $87.621\pm5.523^{\circ}$ & $84.674^{\circ}$ & $-2.947^{\circ}$ & $3.48\%$ & $-0.53$ \\
Fuel $\Delta p$ & $311.267\pm44.963$~kPa & $313.295$~kPa & $+2.028$~kPa & $0.65\%$ & $+0.05$ \\
Ox $\Delta p$ & $253.004\pm11.646$~kPa & $246.505$~kPa & $-6.498$~kPa & $2.64\%$ & $-0.56$ \\
\hline
\multicolumn{6}{@{}l}{\textit{Five-element allocation}} \\
Spray angle & $85.993\pm5.686^{\circ}$ & $87.767^{\circ}$ & $+1.775^{\circ}$ & $2.02\%$ & $+0.31$ \\
Fuel $\Delta p$ & $378.427\pm44.035$~kPa & $378.882$~kPa & $+0.455$~kPa & $0.12\%$ & $+0.01$ \\
Ox $\Delta p$ & $225.462\pm11.031$~kPa & $228.493$~kPa & $+3.031$~kPa & $1.33\%$ & $+0.27$ \\
\hline
\end{tabular}
\end{table*}

\subsubsection{Fluent Combustion-Field Assessment}
\begin{figure*}[!t]
\centering
\includegraphics[width=0.82\textwidth]{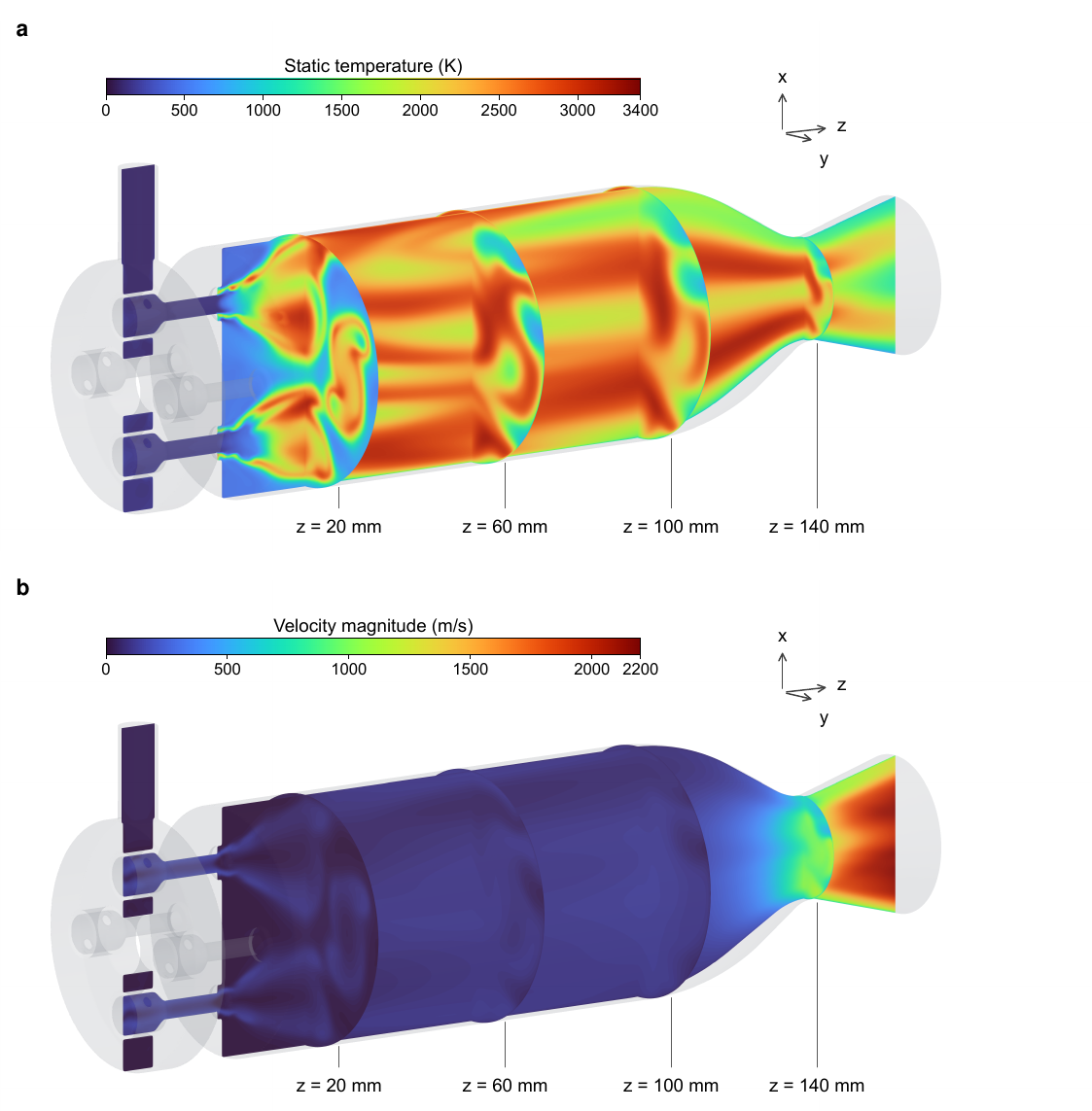}
\caption{Chamber-scale Fluent verification for the four-element configuration. (a) Three-dimensional static-temperature field on a 0--3400~K scale. (b) Corresponding velocity-magnitude field on a $0$--$2200~\mathrm{m\,s^{-1}}$ scale.}
\label{fig:fluent-temperature-field}
\end{figure*}

After passing the injector VOF stage, verification proceeds to Fluent for chamber-combustion and flow-distribution assessment. Fig.~\ref{fig:fluent-temperature-field} shows the representative reacting-flow fields for the four-element configuration. Panel (a) displays the three-dimensional static-temperature field on a 0--3400~K scale, whereas panel (b) shows the corresponding velocity magnitude on a $0$--$2200~\mathrm{m\,s^{-1}}$ scale. Both panels expose the injector-side entry, chamber, convergent section, and nozzle and use sections at $z=20$, 60, 100, and 140~mm to show the axial evolution. The integral comparison for both layouts is reported in Table~\ref{tab:fluent-integral}. The registered Fluent calculations use the reacting-flow setup in Section~\ref{subsec:verification-refinement}.

The computed fields show a continuous and approximately axisymmetric reacting flow through the realized chamber. The temperature field develops a centered high-temperature core with smooth radial and axial transitions from the injector face to the throat and exit, without isolated hot spots or abrupt discontinuities dominating the solution. The velocity field shows the complementary expected pattern: lower velocities in the head-end mixing region, progressive acceleration through the convergent section, and the largest magnitudes near the throat and in the divergent nozzle. The coherent temperature and velocity distributions indicate adequate mixing and orderly expansion, consistent with reported numerical assessments of regeneratively cooled rocket thrust chambers \cite{kim2014multidisciplinary,cutrone2008cfd}. For this preliminary-design study, these field patterns provide the qualitative flow-organization evidence, while the integral quantities in Table~\ref{tab:fluent-integral} provide the quantitative screening check. The ideal-gas oxidizer closure and the effective Fuel spray boundary condition are retained as declared preliminary-model assumptions; their detailed definitions are documented in ~\ref{app:simulation-settings}.

An independent performance-summary workbook was used to compare the reacting-flow outputs with the accepted design state. Both layouts retained $r_{\mathrm{Ox/Fuel}}=1.0901$, a total flow of $0.283~\mathrm{kg\,s^{-1}}$, a throat diameter of $22.8~\mathrm{mm}$, and the same throat area of $4.08281\times10^{-4}~\mathrm{m^2}$, corresponding to deviations of only 0.01\% in mixture ratio and 0.02\% in total flow from the design ledger. These differences describe agreement of the specified operating conditions with the design ledger. Table~\ref{tab:fluent-integral} reports the absolute responses and their deviations from the corresponding design or RocketCEA references. All four integral responses for both layouts remain within the declared $|\delta|<10\%$ preliminary-design criterion; the largest listed deviation is 7.91\%. The four-element layout is consistently closer to the target state, while the five-element snapshot is retained as lower-confidence evidence because of its approximately 2.5\% mass imbalance. These combustion-response results do not complete the Fluent stage: both layouts must also undergo the flow-distribution check before either can enter Abaqus.

\begin{table*}[!t]
\centering
\footnotesize
\setlength{\tabcolsep}{5pt}
\renewcommand{\arraystretch}{1.08}
\caption{Fluent integral performance responses for the four- and five-element layouts.}
\label{tab:fluent-integral}
\begin{tabular}{@{}lccccc@{}}
\hline
\textbf{Metric} & \textbf{Design reference} & \textbf{Four elements} & \textbf{Relative error} & \textbf{Five elements} & \textbf{Relative error} \\
\hline
$\overline{p}_{\mathrm{c}}$ & $1.000$~MPa & $0.9441$~MPa & $-5.59\%$ & $0.9303$~MPa & $-6.97\%$ \\
$F$ & $500$~N & $471.63$~N & $-5.67\%$ & $460.53$~N & $-7.89\%$ \\
$I_{\mathrm{sp}}$ & $180.197$~s & $169.939$~s & $-5.69\%$ & $165.940$~s & $-7.91\%$ \\
$c^*$ & $1403.5~\mathrm{m\,s^{-1}}$ & $1362.1~\mathrm{m\,s^{-1}}$ & $-2.95\%$ & $1342.2~\mathrm{m\,s^{-1}}$ & $-4.37\%$ \\
\hline
\end{tabular}
\end{table*}

\subsubsection{Fluent Flow-Distribution Screening}
\begin{figure*}[!t]
\centering
\includegraphics[width=0.96\textwidth]{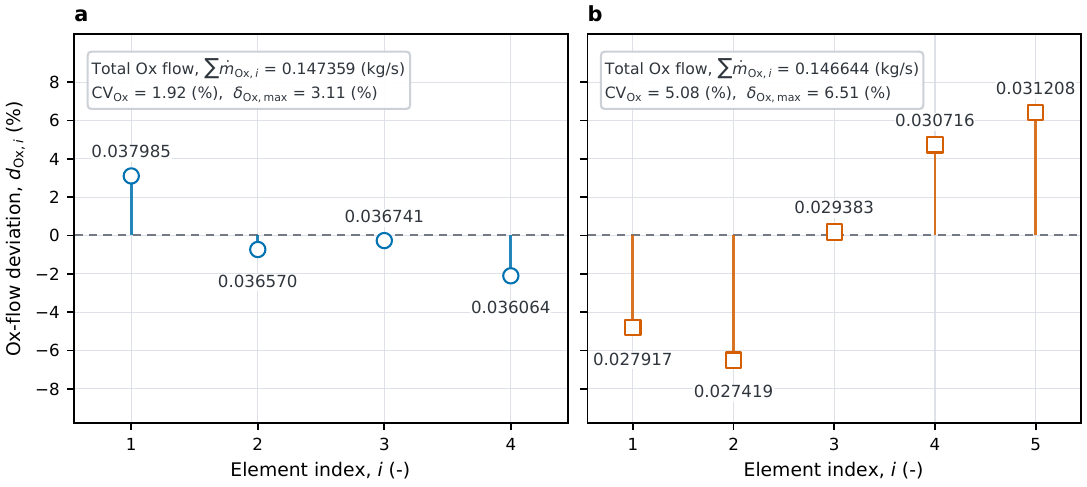}
\caption{Oxidizer-side branch mass-flow distributions under the prescribed inlet flow. (a) Four elements; (b) five elements.}
\label{fig:injector-flow-uniformity}
\end{figure*}

Both layouts reached the flow-distribution check after VOF acceptance and satisfaction of the Fluent combustion-response criterion. Their Ox outlet distributions determine whether each candidate can proceed to Abaqus. The distribution plot is intentionally limited to Ox: the CFD model resolves the Ox feed passages and obtains each branch flow by outlet integration, whereas the Fuel total flow is imposed directly through the spray-injection boundary condition rather than being solved as a separate branching network. The $<1\%$ inlet-to-outlet and pairwise flow checks therefore apply only to the resolved Ox feed. The prescribed Fuel input has no corresponding resolved Fuel branch-flow acceptance test. Fig.~\ref{fig:injector-flow-uniformity} compares the two layouts. Circles in (a) represent four elements and squares in (b) five elements; each marker reports $d_{\mathrm{Ox},i}=100(\dot{m}_{\mathrm{Ox},i}/\overline{\dot{m}}_{\mathrm{Ox}}-1)$ relative to its layout mean, indicated by the dashed zero line. The four-element layout in (a) passed this distribution screen, while the wider spread of the five-element layout in (b) led to its rejection before structural analysis.

The integrated Ox outlet totals, shown rounded in Fig.~\ref{fig:injector-flow-uniformity}(a) and (b), are 0.14736 and 0.14664~kg/s. Relative to the common prescribed Ox inlet of 0.14756~kg/s, their absolute flow-balance deviations are 0.139\% and 0.624\%, both below 1\%. Their mean-normalized pairwise difference is 0.487\%, also below 1\% but distinct from either inlet-to-outlet deviation. These reported balances concern the Ox circuit; the prescribed Fuel flow remains 0.13538~kg/s. With Ox total-flow acceptance established, the tighter distribution in (a) gives a lower coefficient of variation (1.92\% versus 5.08\%) and worst-branch deviation (3.11\% versus 6.51\%). The four-element values, $\mathrm{CV}_{\mathrm{Ox}}=1.92\%$ and $\delta_{\mathrm{Ox},\max}=3.11\%$, both satisfy the respective $5\%$ limits, so this layout advanced to Abaqus. The five-element values, $5.08\%$ and $6.51\%$, both exceed $5\%$; this layout was therefore discarded at the Fluent stage, and no Abaqus strength calculation was performed for it.

\subsubsection{Top-Plate Thermal--Structural Assessment}
\begin{figure*}[!t]
\centering
\includegraphics[width=0.96\textwidth]{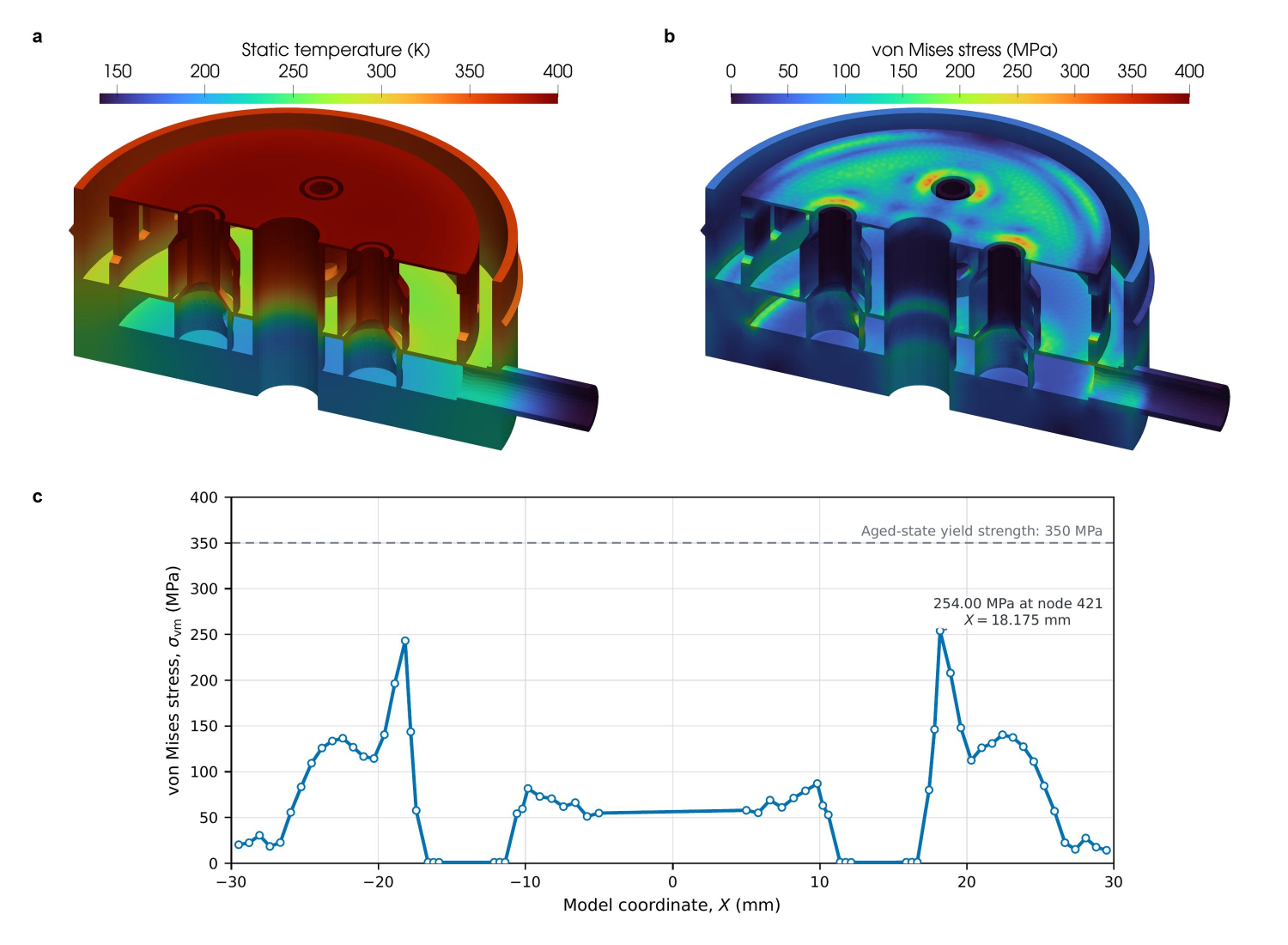}
\caption{Thermal--structural response of the selected four-element injector assembly. (a) Nodal temperature; (b) von Mises stress; (c) averaged stress along the top-plate path.}
\label{fig:abaqus-structural-evidence}
\end{figure*}

Only the four-element layout passed the preceding VOF and Fluent combustion and flow-distribution checks. It therefore entered Abaqus as the sole surviving candidate for the final strength check; the five-element layout had already been discarded. The injector top plate transfers combined pressure and thermal loads around the injector and igniter openings, making it a critical region for structural verification. Fig.~\ref{fig:abaqus-structural-evidence} connects this loading to the four-element candidate assembly's response. The nodal-temperature field in (a), spanning 140.18--396.38~K, supplies the thermal load for the structural calculation in (b), together with the pressure loads. These views locate the coupled response around the openings and surrounding housing. Panel (c) supplies the quantitative stress assessment.

Along the declared top-plate path in Fig.~\ref{fig:abaqus-structural-evidence}(c), the von Mises stress ranges from 0.96 to 254.00~MPa, peaking at $X=18.175~\mathrm{mm}$. Relative to the approximately 350~MPa aged-state CuCrZr yield reference, this gives a path-based stress margin of 1.38, defined as the ratio of the yield reference to the maximum path stress, and satisfies the minimum criterion of 1.2. The stress distribution along the declared top-plate path and around the injector--igniter load-transfer region remains spatially coherent and physically reasonable for the preliminary structural assessment. The four-element assembly passed the prescribed top-plate check, completing the declared preliminary top-plate structural check and supporting its final selection.

The flow-distribution screen eliminated the five-element layout, and the subsequent Abaqus pass established acceptance of the surviving four-element configuration under the declared preliminary structural criterion. The selected assembly combines an operating point of $r_{\mathrm{Ox/Fuel}}=1.09$ and $I_{\mathrm{sp}}=180.20$~s with chamber, throat, and exit radii of 28.0, 11.4, and 17.1~mm, respectively; $L_{\mathrm{c}}=142.2$~mm and $L_{\mathrm{e}}=21.8$~mm. Its 32-channel balanced cooling layout gives $T_{\mathrm{w},\max}=725.3$~K, $T_{\mathrm{out}}=379.8$~K, a pressure loss of 0.221~MPa, and a minimum low-order safety factor of 1.80 under the common material card. Combined with the admitted four-element injector, these outputs form the geometry shown at the start of this section. The progression from the revised operating point to this verified and selected configuration realizes the shared-state continuity described in Section~\ref{subsec:overall-framework}.

\subsubsection{Workflow-Level Refinement and Configuration Reuse}
\label{subsubsec:results-recovery}

The physical checks above end with a selected configuration; the workflow must then show that this state can be revised without silently reusing superseded evidence. The engineering sequence also includes correction of a thermal-profile/selected-geometry binding followed by repeated low-order structural analysis (~\ref{app:agent-execution}). These actions illustrate continuation after a bounded search found no hard-feasible cooling candidate and after an input-binding error was identified. The following controlled tests therefore move from physics-level verification to workflow-level refinement: dependency-aware revision handling and reuse of the geometry interface across nozzle families.

\paragraph{Revision-Aware Workflow Validation.}
\label{subsubsec:revision-aware-validation}

Two configurations are compared on three controlled scenarios from the same frozen design snapshot. The full configuration enables automatic invalidation and revision-aware admission; the comparison disables only these checks while retaining the same baseline, invocation parameters, snapshot-integrity checks, and file-existence and checksum checks. An upstream mixture-ratio change and a structural-only material change exercise the main state-update path; submission of a superseded STL exercises the actual Abaqus adapter's entry checks. Each scenario starts from the same baseline in both configurations, giving six deterministic executions that stop before solver launch.

Expected affected sets were registered before execution, and the replay records missed and excess invalidations, additions to the re-execution plan, and stale-file admission outcomes (~\ref{app:reproducibility}). The full configuration reproduced both expected invalidation sets and blocked the superseded STL. The upstream mixture-ratio change invalidated 24 artifacts across 11 stages and added seven stages to the dependency-ordered re-execution plan, whereas the configuration with revision checks disabled detected none of these newly affected records. The structural-only material change invalidated five artifacts across four stages in the full configuration; the affected stages were already pending when checks were disabled, so newly invalidated records rather than plan differences establish detection. For the superseded STL, the full configuration rejected the inactive file at the Abaqus entry while the comparison configuration accepted it at preflight.

\paragraph{Nozzle-Configuration Reuse.}

Configuration reuse was tested by changing only the nozzle-shape branch while holding thrust (500~N), chamber pressure (1.0~MPa), propellants (LOX/RP-1), mixture ratio (1.05), contraction ratio (6), 0.1-mm parameter quantization, and a common cooling/wall configuration fixed. The TOC, $15^{\circ}$ conical, and $30^{\circ}/10^{\circ}$ parabolic-bell profiles all retained the injector-face, chamber-end, throat, and exit anchors with strictly increasing axial coordinates. Maximum radial reconstruction error was at most 0.1~mm; profile-wise maximum interpolation errors were 0.1000~mm for TOC, 0.09828~mm for the conical contour, and 0.1000~mm for the parabolic bell. All six solid/fluid STLs passed binary-format validation, the expected log markers were present, and 15 generator regression tests passed. Profile discretization and triangle counts are retained in ~\ref{app:embodiment-settings}.

\begin{figure*}[!t]
\centering
\includegraphics[width=\textwidth]{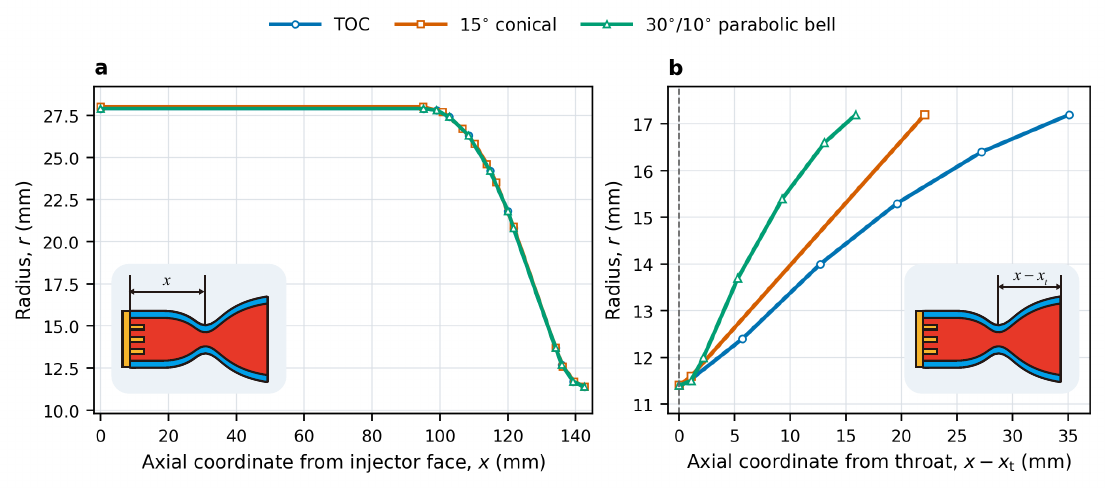}
\caption{Contour comparison for the three nozzle profiles. (a) Chamber and convergent sections; (b) divergent sections.}
\label{fig:nozzle-configuration-adaptation}
\end{figure*}

Fig.~\ref{fig:nozzle-configuration-adaptation} separates the contour comparison at the throat. Panel (a) shows the nearly coincident chamber and convergent profiles from the injector face to the throat, using the axial coordinate $x$. Panel (b) shows only the divergent sections, with each throat reset to $x-x_{\mathrm{t}}=0$. The insets illustrate these coordinate definitions on a schematic thrust chamber. The parabolic bell reaches its exit over the shortest downstream distance, followed by the conical contour and TOC. Including the upstream sections, their complete profile lengths are 158.5, 164.8, and 177.7~mm, respectively. Open markers identify the retained sparse coordinates. Together with the common validation results, the profiles demonstrate reuse of the geometry interface across nozzle families at the declared tolerance.

Together, the controlled tests show how the shared-state and geometry interfaces support the executed engineering sequence: revision checks identify dependent work and reject inactive inputs, while the same geometry contract accommodates different contour families. These mechanism-level results complement the representative design's physical assessments and selection outcome.

\section{Discussion}
\label{sec:discussion}

\subsection{Long-Horizon Workflow Execution}
\label{subsec:trustworthiness}

Long-horizon execution here means that the Agent drives an extended sequence of dependent engineering tasks from requirements to a reviewed preliminary design. The recorded plan retained unresolved work across stages and kept geometry generation pending until subsystem admission was resolved (~\ref{app:agent-execution}). This makes progression an engineering decision: a completed calculation enables further work only when its outputs satisfy the relevant conditions. Conditional human approval also supports continuation. The cooling failure activated a previously authorized rich-side search without granting unrestricted changes to chamber pressure or safety limits.

Embodiment refinement and candidate verification have distinct failure handling. Failure to find a feasible cooling candidate within the configured search space and budget prompted an authorized operating-point redesign, while the thermal-profile/selected-geometry mismatch required input rebinding and a repeated low-order structural calculation during embodiment. Once a realized candidate enters the linear VOF--Fluent--Abaqus verification sequence, any failed check discards that candidate. The five-element layout illustrates this rule: its Fluent flow-distribution failure terminated the sequence before Abaqus. The controlled revision tests in Section~\ref{subsubsec:revision-aware-validation} address a third problem: preventing downstream use of evidence superseded by a design change. These tests support the invalidation and solver-entry mechanism separately from the historical execution record. Across these forms of continuation, one coordinator retains the global plan and decision ownership, deterministic programs evaluate candidates and enforce numerical gates, and engineers authorize consequential changes and review the outcome.

\subsection{Comparative Engineering Value}
\label{subsec:comparative-value}

The recorded initial sizing and subsystem-screening segment gives a concrete example of Agent coordination. One user confirmation established the operating defaults and conditionally authorized a rich-side search at fixed chamber pressure. Before the first consolidated report, no further user message was recorded: the Agent continued after none of the 74 initial cooling evaluations yielded a hard-feasible candidate, refined the viable region, and invoked the confirmation run at $r_{\mathrm{Ox/Fuel}}=1.09$ that returned eight hard-feasible candidates. It also inspected a ranked cooling artifact after a low-order structural solver had returned a completed pass, identified an inconsistency in its own earlier thermal-profile/candidate binding, corrected the binding, and repeated the calculation. This action required interpreting the relationship between artifacts beyond the individual solver's pass signal. The repair concerned a historical candidate state, separate from the final balanced cooling selection. The initial segment ended with geometry generation pending because injector admission remained unresolved; later work included further author decisions, geometry generation, and multiphysics verification.

RocketAgent organizes the decisions and handoffs between computational tasks. Fixed scripts and multidisciplinary optimization provide reproducible calculations once their inputs, formulation, and execution order are specified \cite{martins2013multidisciplinary}. With these capabilities available as skills, the Agent carries the accepted requirements forward, interprets returned results, selects the next eligible action, and maintains the associated input and artifact bindings. This division of work makes existing engineering tools accessible through a continuing design task while preserving their responsibility for physical predictions and numerical checks.

The execution record separates human decisions, Agent actions, and deterministic computations. The subsequent six revision-governance replays add controlled evidence for dependency invalidation and solver-entry blocking when an input becomes superseded. Together, the historical trace and the replays show how a design task can continue through permitted revisions while retaining unresolved work and evidence dependencies. The historical trace documents coordination in use, while the replays establish the behavior of specific supporting mechanisms.

Following the task-level accounting used by Chen and Bao \cite{chen2026uhpcmultiagent}, we made a corresponding direct estimate for the present workflow. The retained task log contains approximately $1.53\times10^{5}$ tokens for the representative design run; using the conservative API-equivalent basis described in ~\ref{app:agent-execution}, this corresponds to about 33~CNY in LLM token cost. A manual baseline of approximately 3~h at 70~CNY/h is 210~CNY, so the direct comparison is about 210 versus 33~CNY before solver, software, framework, and residual-supervision costs. These values are estimates based on the recorded token count and stated pricing assumptions, rather than measured savings or a controlled comparison; the detailed accounting is given in ~\ref{app:agent-execution}.

\subsection{Framework Evolvability}
\label{subsec:evolvability}

RocketAgent supports extension at the configuration, computational-capability, and knowledge levels. Alternative injector counts and nozzle contours demonstrate reuse of common interfaces with configuration-specific outputs. Materials, propellants, operating conditions, and cooling architectures can likewise be represented through backend definitions, applicability conditions, and engineering gates. Extending these definitions requires evidence appropriate to the new configuration. The capability--skill--adapter separation in Section~\ref{subsec:overall-framework} extends this reuse to replacement solvers or higher-fidelity analyses through declared inputs, outputs, units, validators, and dependency effects. Knowledge evolves through the review process in Section~\ref{subsec:verification-refinement}: completed revisions and feedback provide provenance-linked experience, from which reviewed, applicable guidance can be reused. Together, these extension paths provide a basis for broader product coverage.

\subsection{Limitations and Future Work}
\label{subsec:limitations-future-work}

The representative workflow and module tests support simulation-based preliminary design. They do not establish ignition/shutdown behavior, combustion stability, cyclic life, feed-system dynamics, or manufacturing qualification. Material characterization, cold-flow testing, and hot-fire evidence remain necessary for component qualification. Future work will connect experimental evidence, extend configuration and analysis coverage, and examine whether reviewed experience improves subsequent design work across repeated workflows.

\section{Conclusions}
\label{sec:conclusions}

RocketAgent enables a Coding Agent to drive an extended multidisciplinary workflow for preliminary thrust-chamber design. To the best of the authors' knowledge, it is the first knowledge-constrained single-coordinator, multi-skill architecture explicitly directed at preliminary design of a complex industrial product such as a liquid-rocket thrust chamber. The Agent maintains the overall plan and coordinates skills from requirements through subsystem design, geometry realization, and multiphysics evaluation. The KG and Design IR support cross-stage execution, while revision-aware invalidation and solver-entry admission manage dependent work when the design changes.

The representative design demonstrates closed-loop refinement and configuration selection. Changing to $r_{\mathrm{Ox/Fuel}}=1.09$ moved the cooling search from 0/74 to 8/70 hard-feasible candidates. Injector screening linked physical-flow allocations to water-reference surrogate predictions, assessed using 118 held-out and 140 independent post-selection VOF evaluations. Verification proceeded linearly through VOF, Fluent combustion and flow-distribution screening, and Abaqus strength checking, with immediate rejection on failure. Both layouts met the reported combustion-response and Ox total-flow checks. Flow-distribution acceptance required both the Ox-flow coefficient of variation and maximum branch deviation to be no greater than $5\%$. The four-element layout satisfied both limits, with values of $1.92\%$ and $3.11\%$, while the five-element layout exceeded both, with values of $5.08\%$ and $6.51\%$. The five-element layout was therefore discarded before Abaqus. The four-element layout advanced to the final top-plate strength check, passed it, and was accepted as the selected configuration.

Two configurations and three controlled scenarios tested revision handling in the main workflow. The full configuration reproduced the expected upstream and local invalidation sets and rejected a superseded STL at solver entry. Disabling revision checks removed these responses.

RocketAgent contributes to engineering informatics by sustaining Agent coordination across a long engineering task chain: selecting permitted continuations from analysis feedback, maintaining input and artifact associations, and keeping unresolved prerequisites visible until work can advance. Existing engineering programs provide the numerical capabilities, while auditable state management and explicit decision ownership connect them into a continuing design process. The simulation-based results establish a foundation for subsequent experimental qualification.

\appendix
\section{Evaluation Scope and Reproducibility}
\label{app:reproducibility}
The representative evaluation uses 500~N, sea-level static operation ($P_{\mathrm{amb}}=1~\mathrm{atm}$), $P_{\mathrm{c}}=1.0~\mathrm{MPa}$, LOX/RP-1 at the accepted $r_{\mathrm{Ox/Fuel}}=1.09$, a $15^{\circ}$ conical nozzle, regenerative cooling, and four- and five-element injector layouts. Configured search bounds are $P_{\mathrm{c}}=0.4$--1.20~MPa and RocketCEA $r_{\mathrm{Ox/Fuel}}=0.4$--$4.0$. By comparison, realized RPA artifacts span $r_{\mathrm{Ox/Fuel}}=0.9$--2.6596, with thermally feasible recorded samples at approximately 1.0--1.81. These configured and sampled ranges are not universal feasibility or qualification limits. LOX enters at 90.18~K and the fuel coolant at 293~K and 2.5~MPa.

Both the cooling structural screen and Abaqus use the input card's constant isotropic CuCrZr properties: $k=320~\mathrm{W/(m\,K)}$, $E=128~\mathrm{GPa}$, $\nu=0.34$, $\alpha=1.70\times10^{-5}~\mathrm{K^{-1}}$, and stress-free $T_{\mathrm{ref}}=293.15~\mathrm{K}$, without a temperature-dependent property table or plasticity model. Component strength criteria remain distinct. Cooling admission requires $\mathrm{SF}\geq1.5$ relative to a conservative 120-MPa screening allowable, which is a component-level policy rather than a CuCrZr yield property. Evaluating the selected balanced cooling design (\texttt{case\_063}) with the common $E=128$~GPa material card gives a maximum combined screening stress of 66.82~MPa and a minimum safety factor of 1.80, reported to two decimal places. The evaluation preserves the geometry and retained thermal fields of each candidate, and the eight hard-feasible candidates remain feasible. The selected design is \texttt{case\_063}; the temperature-priority alternative \texttt{case\_053} has a minimum safety factor of 1.95 under the common material card and is not the selected cooling configuration. The Abaqus calculation uses an approximately 350-MPa aged-state yield reference and the same nominal top-plate path metric. Its path maximum of 254.00~MPa gives a path-based stress margin of 1.38, defined as the ratio of the yield reference to the maximum path stress, exceeding the minimum criterion of 1.2. Local contour extrema remain diagnostic and do not replace the declared path metric.

The remaining cooling hard gates require $T_{\mathrm{w}}\leq850~\mathrm{K}$, single-phase coolant, compliance with the pressure-loss budget, RP-1 bulk temperature $\leq399~\mathrm{K}$, cooling-side wall temperature $\leq728~\mathrm{K}$, and channel width and depth each $\geq1.0~\mathrm{mm}$ after 0.1-mm quantization. Auxiliary robust targets of $T_{\mathrm{w}}\leq820~\mathrm{K}$, pressure loss $\leq0.8~\mathrm{MPa}$, and $\mathrm{SF}\geq1.7$ are assessed separately from hard admission. At configuration level, the $<1\%$ flow criterion applies only to the resolved Ox feed. For each layout, the absolute difference between the sum of integrated Ox branch outlet flows and the prescribed Ox inlet flow is normalized by that inlet flow. The absolute difference between the two layouts' Ox outlet totals is normalized by their arithmetic mean. Each inlet-normalized deviation and the pairwise difference must be below 1\%. Fuel is introduced at a prescribed spray-injection rate rather than through a separately resolved branching network, so this test does not impose a Fuel or whole-combustor $<1\%$ acceptance requirement. The Ox-channel check and the whole-combustor numerical conservation diagnostics are distinct. Within-layout flow distribution is a separate pass-or-reject check within the Fluent stage, requiring both $\mathrm{CV}_{\mathrm{Ox}}\leq5\%$ and $\delta_{\mathrm{Ox},\max}\leq5\%$. Exceeding either limit discards the candidate. Passing the Ox total-flow checks does not waive this requirement; only candidates passing both the combustion and flow-distribution checks proceed to Abaqus.

The reacting-flow combustion screen uses the same relative-deviation definition as the main text and requires $|\delta|<10\%$ for chamber pressure, thrust, specific impulse, and $c^*$ relative to their accepted-design or RocketCEA references. This is a preliminary-design consistency criterion for comparing realized configurations; it does not establish high-fidelity solver accuracy, real-fluid model validity, or hardware qualification.

\begin{figure*}[!t]
\centering
\includegraphics[width=0.62\textwidth]{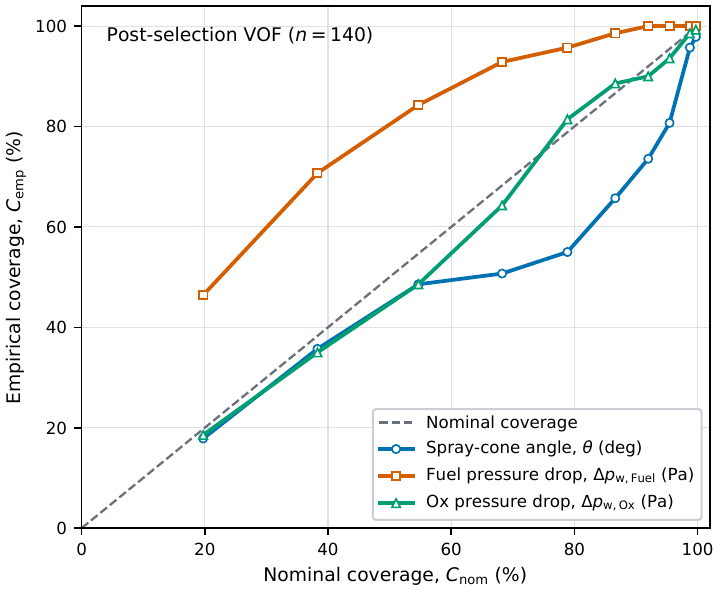}
\caption{Calibration diagnostic for the 140 independent post-selection water-reference VOF evaluations. The diagonal denotes equality between empirical and nominal central-interval coverage. Curves below it indicate under-coverage and curves above it indicate conservative intervals. The candidates are selected by optimization; the curves describe this search-conditioned population, not a random held-out test or a guarantee for real-propellant operation.}
\label{fig:gpr-coverage}
\end{figure*}

\paragraph{Predictive uncertainty.}
Fig.~\ref{fig:gpr-coverage} compares empirical central-interval coverage with Gaussian nominal coverage $2\Phi(k)-1$ for the 140 post-selection evaluations. Relative to the equality diagonal, the spray-angle curve lies below nominal coverage across most levels, whereas the Fuel-pressure curve lies above it; the Ox-pressure curve follows the diagonal more closely. The three curves therefore distinguish underestimated spray-angle uncertainty from conservative Fuel-pressure intervals, extending the two coverage levels in Table~\ref{tab:gpr-calibration}. The Gaussian negative log predictive density in that table is $\mathrm{NLPD}_j=(2n)^{-1}\sum_i[\log(2\pi\sigma_{ij}^{2})+z_{ij}^{2}]$. No recalibration is fitted on these labels, and the resulting coverage applies to the evaluated benchmark rather than configuration-specific inference.

\paragraph{KG retrieval settings.}
The following settings implement the retrieval and applicability procedure in Section~\ref{subsec:kg-conceptual-design}. Stage and the internal design-record identifier set the retrieval context. Prior-design scores weight thrust, chamber pressure, Ox, and Fuel by 0.20 each, and nozzle configuration and cooling requirement by 0.10 each. Thrust and pressure use N and MPa. Positive numerical fields use $s_{\mathrm{num}}(a,b)=\max\{0,1-|\ln(a/b)|/\ln(10)\}$. Categorical fields use lowercase exact or substring matching after removing hyphens and underscores. Missing or nonpositive numeric values and missing categories contribute zero without weight renormalization; ties preserve index order.

Default retrieval returns up to three positive-scoring prior designs, four positively ranked RPA examples, and six stage-compatible method cards. RPA ranking uses required analysis features, ChannelWallSegment/thermal capability, propellant compatibility, pressure proximity, nozzle features, and example class. Reviewed facts are filtered by stage and topic aliases: all matches remain in the linked response, with at most five displayed in the compact trace. Relation extraction retains up to 20 edges from each of six relation families: design record, stage, method, requirement--function, conflict, and experience. The 120-assignment limit permits repeated edges across families. Rule-card JSON pointers specify the scope, unit, range, state, and artifact predicates used for applicability checks.

The worked knowledge trace is retained in the execution record as a provenance-linked sequence of initial concept and cooling advice, applicability checks, adoption or rejection decisions, and subsequent calculations. It records historical case 006 as a bounded precedent, the authorized continuation after the initial cooling search, and the distinction between the historical recommendation of $r_{\mathrm{Ox/Fuel}}=1.65$ and the downstream calculated value of 1.09. Later synchronized advisor outputs are not used to reconstruct the initial recommendation. This record preserves the temporal boundary between retrieved guidance and current numerical evidence without exposing implementation-specific file paths in the method description.

\paragraph{Code and artifact availability.}
The frozen evidence package contains the Design-IR and workflow schemas, solver inputs and configuration files, scripts, geometry and mesh artifacts, native and processed outputs, validation reports, decision records, manifests, and checksums supporting the reported results. The code and data supporting the findings of this study are available from the corresponding author upon reasonable request.

\paragraph{Revision-governance replay records.}
The replay package retains the frozen baseline, two configuration files differing only in the revision-check switch, preregistered expected sets, six execution traces, and set-by-set comparisons. Expected sets are specified in a separate protocol file, hashed before execution and loaded for comparison only after all traces are closed. They are protocol-preregistered without a claim of independent human sign-off. The upstream perturbation changes the selected mixture ratio from 1.09 to 1.095. The structural perturbation changes an isolated analysis card from 120 to 125~GPa; these workflow-test values do not replace the shared 128-GPa CuCrZr definition used in the physical analyses.

The superseded-file check uses a five-element geometry branch in a benchmark-local copy of the ledger without altering the selected four-element design. This preflight-only governance replay is separate from physical candidate verification: it tests the adapter entry contract, does not reinstate the layout rejected at the Fluent flow-distribution check, and performs no strength analysis. The actual Abaqus adapter returns exit code 3 for the full configuration and 0 for the comparison configuration, with neither invocation creating a solver deck or launching a solve. The baseline already contains pending Abaqus, final-decision, validation, and knowledge-synchronization stages because required outputs are absent. This common backlog explains why the local-change plans are identical despite different invalidation records. The six executions are deterministic governance replays, not independent LLM trials, and report no avoided-solver cost.

\section{Agent Configuration and Execution Records}
\label{app:agent-execution}

\paragraph{Runtime and session context.}
The representative design was developed in Codex Desktop on a local Windows workspace. The retained task log records client identifier 0.147.0-alpha.6.6, OpenAI model gpt-5.6-sol, and low reasoning-effort settings in different recorded turns. Work continued in the same persistent task across multiple dates, including context-compaction events. Engineering stages were therefore not independent LLM trials. The log records neither temperature nor an immutable backend model release, and these settings are not inferred from application defaults.

\paragraph{Instructions and decision ownership.}
Author-controlled instructions comprise the engineering request, project operating rules, the staged workflow, and task-specific skill documents. The Agent loads the relevant skill before invoking its scripts and maintains an ordered task plan with pending, in-progress, and completed states. It selects eligible actions, binds Design IR inputs, interprets returned evidence, and proposes revisions within the approved scope. Deterministic programs implement numerical search, objective evaluation, engineering thresholds, and revision propagation. One lead Agent owns the global plan, with bounded auxiliary agents recorded for initial retrieval and readiness checks. Code and configuration editing are available, and requested adapter changes are logged with their checks. Human authorization governs changes to requirements, accepted models, and safety limits. The single-coordinator description thus refers to plan and decision ownership.

\paragraph{Observed planning and execution.}
The initial request specified a 500-N LOX/RP-1 design. Before computation, the Agent proposed operating conditions and obtained user confirmation, including conditional permission for a rich-side search at fixed chamber pressure if thermal screening failed. The recorded seven-step plan covered conceptual acceptance, performance, geometry, injector screening, cooling and low-order structure, STL generation, and final checking and reporting. Plan updates recorded progression and kept geometry generation pending when subsystem admission was unresolved. The performance call used the accepted 500~N thrust and 1.0~MPa chamber pressure, with sea-level ambient pressure converted to the backend unit.

After the maximum-specific-impulse cooling search returned no hard-feasible candidate, the Agent invoked the authorized rich-side search and refined the feasible region. The confirmation run at $r_{\mathrm{Ox/Fuel}}=1.09$ used the RocketCEA-derived physical Fuel flow of 0.135380~kg/s and retained eight hard-feasible candidates. The initial segment, from the design request through its first consolidated report, contains the initial request and one user confirmation, with no further user message before the report. During this segment, the human confirmed the operating defaults and the permitted rich-side continuation; the Agent established and maintained the staged plan, selected eligible actions, corrected an inconsistent thermal-profile/candidate binding after inspecting the returned artifacts, and kept STL generation pending when injector applicability remained unresolved. Deterministic RocketCEA, RPA, and low-order calculations consumed the accepted inputs, while declared surrogate-domain checks constrained injector admission. Later, a human-requested governance replay used the coordination records as workflow context; it did not constitute a new LLM trial, and its revision checks invalidated dependent records and rejected a superseded STL before solver launch. The initial segment covers performance, quantized contour, cooling, low-order structure, and injector applicability screening; formal PicoGK STL generation was still pending.

\paragraph{Illustrative labor and token-cost comparison.}
For a direct comparison, let $C_{\mathrm{manual}}=rH_{\mathrm{manual}}$ denote manual labor cost and let $C_{\mathrm{token}}=p_{\mathrm{token}}N_{\mathrm{token}}$ denote the LLM token cost. The retained task log for the representative 500~N design run records approximately $N_{\mathrm{token}}=1.53\times10^{5}$ total tokens. Because the log does not separate billable input and output tokens, we use a conservative API-equivalent upper-bound rate of $p_{\mathrm{token}}=30$~USD per million tokens and an assumed exchange rate of 7.2~CNY/USD. This gives
\begin{equation}
\begin{aligned}
C_{\mathrm{manual}}&=(70~\mathrm{CNY/h})(3~\mathrm{h})\simeq210~\mathrm{CNY},\\
C_{\mathrm{token}}&=(30~\mathrm{USD}/10^6~\mathrm{tokens})(1.53\times10^5~\mathrm{tokens})(7.2~\mathrm{CNY/USD})\simeq33~\mathrm{CNY},\\
\Delta C_{\mathrm{direct}}&=C_{\mathrm{manual}}-C_{\mathrm{token}}\simeq177~\mathrm{CNY}.
\end{aligned}
\label{eq:agent-labor-cost}
\end{equation}
The direct token-only estimate is therefore approximately 16\% of the manual baseline, or an 84\% lower direct cost. This comparison excludes unattended solver time, software licensing, framework development and maintenance, and any residual human supervision; it is an API-equivalent scenario based on the retained log rather than an invoice or a controlled measurement of total project cost.

\section{Numerical Simulation Settings}
\label{app:simulation-settings}

\paragraph{Injector cold-flow VOF.}
The snappyHexMesh grid used a 0.5-mm global cell size and 0.125-mm local refinement in the swirl chamber and downstream atomization region, giving approximately $5.5\times10^6$ cells per configuration. Inlet mass flow, outlet pressure, and no-slip walls defined the boundaries. DeepFlame's OpenFOAM-derived interFoam workflow used SST $k$--$\omega$, PIMPLE coupling, and a fixed $10^{-5}$-s time step. Both circuits used water at the mapped flows in Eq.~\ref{eq:injector-water-equivalent-flow}. Extracted quantities comprised spray angle, circuit pressure drops, and phase/velocity fields. Both allocation-specific calculations use the same square-root density-ratio mapping.

The four- and five-element endpoint comparisons use ten samples from 0.091 to 0.100~s at 0.001-s intervals. Spray-cone angles are obtained from fitted liquid-sheet midlines on the root-connected $\alpha_{\mathrm{water}}=0.25$ contour. The four-element samples at 0.097, 0.098, and 0.099~s and the five-element samples at 0.093 and 0.099~s carry the \texttt{MULTIPLE\_INTERSECTIONS} flag, indicating more than two intersections between the extracted liquid contour and a measurement line. This identifies ambiguity in angle and liquid-film-thickness extraction; it does not automatically identify solver failure. The primary comparisons retain all ten samples in each fixed window and report the quality flags. The unflagged subsets are separate sensitivity analyses: seven samples give $83.396^{\circ}$ for the four-element allocation, and eight samples give $87.368^{\circ}$ for the five-element allocation. Their relative deviations from the respective GPR means are approximately 5.1\% and 1.57\%, compared with 3.48\% and 2.02\% for the full windows. Each deviation uses its own VOF mean as denominator. The full-window deviations are the values used for the $5\%$ VOF gate; the unflagged subset is reported as a frame-selection sensitivity analysis and does not replace the declared endpoint assessment.

Circuit pressure drops are the inlet-patch area-averaged static pressure minus the outlet-patch area-averaged static pressure at each sampled time, averaged over the same 0.091--0.100-s window. The pressure means in Table~\ref{tab:vof-endpoint} use all ten samples; the angle-extraction flags do not remove pressure samples. The pressure means are taken from the retained flow-and-pressure summary, while the spray-angle means and flag sensitivity are recomputed from the per-frame records.

\paragraph{Reacting chamber flow.}
The Fluent mesh was generated with snappyHexMesh at 0.5-mm background resolution and 0.25-mm refinement in the chamber head, totaling approximately $7.9\times10^6$ cells. The steady RANS calculation used SST $k$--$\omega$ and a partially premixed FGM model with compressibility correction. Kerosene was represented by $n$-decane with the ten-step reduced Choi mechanism. Hollow-cone fuel injection used $10~\mathrm{m\,s^{-1}}$ velocity, 400~K initial droplet temperature, a $30^{\circ}$ half-cone angle, and a Rosin--Rammler size distribution. The 400~K value is a combustor-model initialization condition, not the calculated cooling outlet temperature. Pressure--velocity coupling used the Coupled algorithm; final spatial discretization was second-order upwind. The Ox feed passages were resolved and their outlet fluxes were integrated branch by branch, whereas the Fuel total flow and spray distribution were prescribed through the injection boundary condition; consequently, the $<1\%$ channel-flow acceptance criterion and the resolved branch-uniformity assessment in the main text apply only to Ox. The prescribed Fuel injection rate specifies a model input; it does not constitute an independently calculated Fuel branch-flow or whole-combustor mass-conservation result.

The injector-angle conventions are distinct. The approximately $80^{\circ}$ quantity reported by the water-reference GPR/DeepFlame analysis is a full spray-cone angle obtained from the two fitted liquid-sheet half-angles, whereas the Fluent model prescribes a $30^{\circ}$ Fuel hollow-cone half-angle. The latter is an effective numerical boundary condition selected to maintain a representative Fuel--Ox overlap under the ideal-gas Ox closure; it is not a conversion of, or a quantitative prediction of, the GPR/VOF angle. Spray geometry also depends on momentum-flux ratio, chamber pressure, and thermodynamic state, so water-reference angles are not transferred directly to the reacting LOX/RP-1 calculation \cite{bee2026characterization}.

\paragraph{Coupled top-plate analysis.}
The selected Abaqus assembly includes the top plate, four peripheral injector passages, central igniter passage, housing transitions, and side port. Its common thermal/structural quadratic-tetrahedral mesh contains 266,393 nodes and 161,274 elements. The steady thermal stage used DC3D10 elements, with film conditions on the actual Fuel- and Ox-wetted surfaces at bulk temperatures of 400 and 90~K, respectively; both heat-transfer coefficients were $5000~\mathrm{W\,m^{-2}\,K^{-1}}$. The thermal solution completed in six increments. Final nodal temperatures were transferred to the matching C3D10 structural mesh; 1.2~MPa pressure was applied to both wetted-surface groups. In the structural calculation, global $U_3$ was constrained on the lower Set-1 interface, the upper Set-2 interface was left unconstrained, and two minimal in-plane constraints were introduced to suppress rigid-body motion. A geometrically linear Abaqus/Standard step combined pressure and thermal strain using the common isotropic CuCrZr card in ~\ref{app:reproducibility}. The analysis considered only the four-element assembly, which passed the preceding VOF and Fluent checks. The five-element layout was discarded for unacceptable flow distribution at the Fluent stage and therefore never entered Abaqus. The Abaqus pressure and film conditions define a separate preliminary structural load case and should not be interpreted as a direct reconstruction of the full regenerative-cooling pressure and temperature fields.

Integration-point stresses were extrapolated to element nodes and arithmetically averaged over adjoining elements, with a 75\% nodal-averaging threshold for contour display. The quantitative top-plate path retained 68 physical nodes in $Y=0\pm0.05$~mm at $Z=-0.2$~mm, over $-29.5\leq X\leq29.5$~mm. The reported maximum and margin use those nodal values; connecting lines are only visual guides. Temperature, heat flux, temperature gradient, stress, displacement, and reaction-force outputs were retained as applicable.

\paragraph{Mesh-independence verification.}
Mesh-independence verification was completed on the high-performance computing platform used for the simulations. The mesh specifications given above describe the meshes used for the reported results. Detailed mesh-comparison records are retained on that platform and are available from the corresponding author upon reasonable request.

\paragraph{Evidence checks.}
Each analysis retains its geometry and operating-state bindings, native outputs, extracted quantities, and disposition. Applicable numerical checks cover residuals, conservation, discretization and iteration stability, mesh quality, and physical consistency. Preparation-only, incomplete, diverged, and failed runs remain in the execution history $L_k$ and do not receive accepted status or replace the active evidence bindings in $S_k$. Candidate verification follows VOF, Fluent combustion and flow-distribution checks, and Abaqus strength checking in sequence; a failed check rejects the candidate and prevents execution of the remaining stages.

\section{Implementation and Search Details}
\label{app:implementation-details}

\subsection{Shared Records and Execution Interfaces}
\begin{figure*}[!t]
\centering
\includegraphics[width=\textwidth]{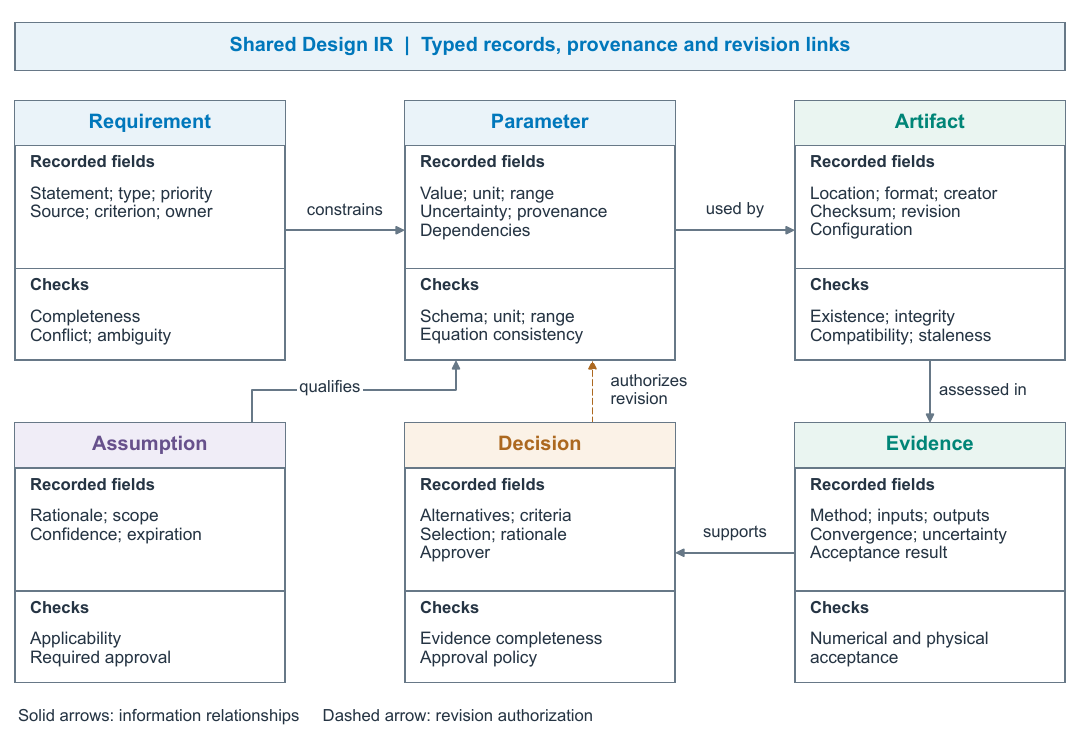}
\caption{Shared Design IR record relationships. Boxes separate fields from usage checks; solid arrows indicate information links, and the dashed arrow denotes authorization of a parameter revision subject to the state-transition contract.}
\label{fig:design-ir-records}
\end{figure*}

Fig.~\ref{fig:design-ir-records} expands the shared state introduced in Section~\ref{subsec:overall-framework}. Across the upper row, requirements constrain parameters, which in turn supply inputs to artifacts. The lower return path connects an artifact to its evidence and then to a decision; the dashed arrow authorizes a parameter revision, closing the recorded design loop. Assumptions separately qualify parameter use. This arrangement retains both the engineering dependency and the reason for changing it, allowing the revision mechanism to trace affected outputs without discarding unrelated records.

\begin{figure*}[!t]
\centering
\includegraphics[width=\textwidth]{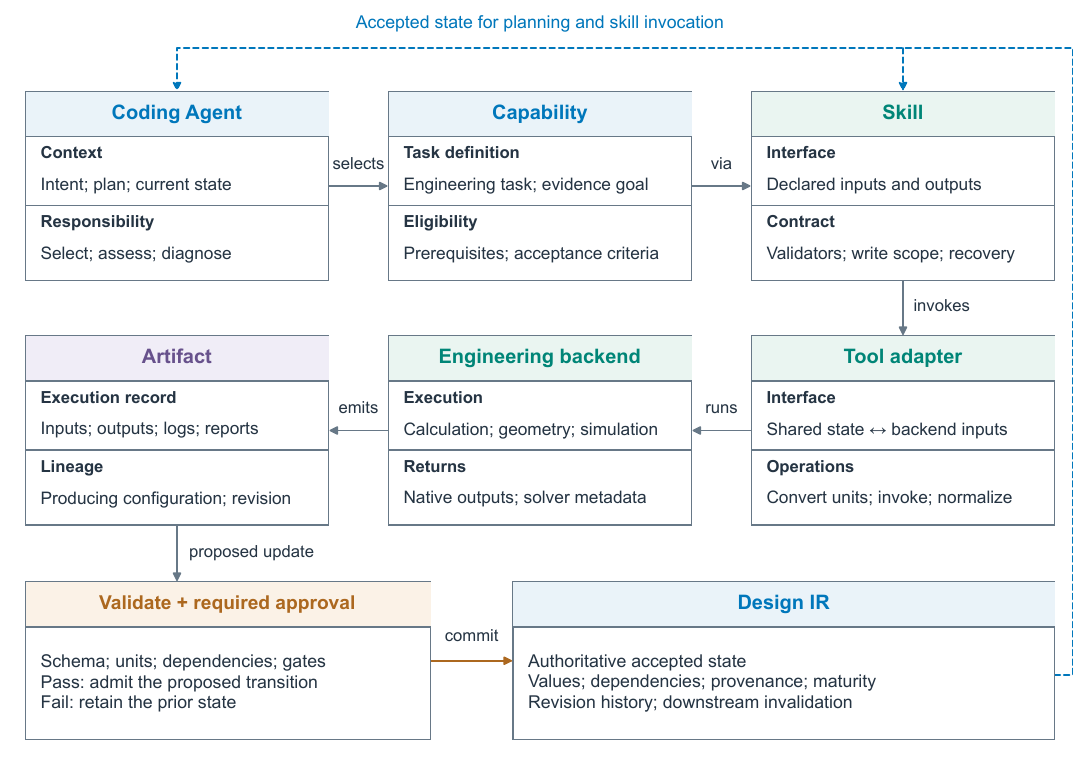}
\caption{Contract-bound execution: capability selection, skill invocation, adapter/backend execution, and checked state commitment. Dashed arrows show accepted-state reads for subsequent planning and execution.}
\label{fig:execution-contract}
\end{figure*}

Fig.~\ref{fig:execution-contract} shows how an individual task updates those records. The forward path connects Agent capability selection to a skill, its adapter, and the engineering backend. The returned artifact then enters validation and any required approval before commitment to the Design IR. The dashed blue return paths make that accepted state available to subsequent planning and skill invocation. Separating artifact production from state commitment ensures that the next task reads a checked result; a failed validation retains the prior accepted snapshot $S_k$ while adding its outputs, failed checks, and disposition to execution history $L_k$. Within the linear verification sequence, the recorded rejection prevents the current candidate from advancing to any remaining solver stage.

\subsection{Embodiment-Design Settings}
\label{app:embodiment-settings}

\paragraph{Performance and contour.}
The evaluated RocketCEA calculation used \texttt{estimate\_Ambient\_Isp} with equilibrium expansion (frozen=0 and frozenAtThroat=0), $g_0=9.80665~\mathrm{m\,s^{-2}}$, and sea-level exit-pressure matching when no expansion ratio was prescribed. An inclusive grid and bounded golden-section refinement retained endpoints over $r_{\mathrm{Ox/Fuel}}=0.4$--$4.0$. Optional prior-design performance bounds required compatible nozzle mode, propellants, chamber pressure, and thrust range or engine class. Without such evidence, the calculated optimum remained the design value. Pressure was converted from the MPa interface convention to the backend convention.

The RPA geometry output comprised $R_{\mathrm{c}}$, $L_{\mathrm{cyl}}$, $L_{\mathrm{c}}$, $\beta$, $R_2$, $R_1$, $R_{\mathrm{t}}$, $R_{\mathrm{n}}$, $\theta_{\mathrm{n}}$, $\theta_{\mathrm{e}}$, $R_{\mathrm{e}}$, $L_{\mathrm{e}}$, exit-area ratio, mass flow, and profile coordinates. Here, $\beta$ is the contraction angle and $\theta_{\mathrm{n}}$, $\theta_{\mathrm{e}}$ are the initial and exit nozzle angles. Conical and TOC profiles were extracted from RPA. Parabolic-bell reconstruction combined RPA sizing, throat rounding, and a tangent-matched quadratic B\'ezier curve. The evaluated configurations used $A_{\mathrm{c}}/A_{\mathrm{t}}=6$, a $15^{\circ}$ cone, and a $30^{\circ}/10^{\circ}$ bell. Dimensions were quantized to $0.1~\mathrm{mm}$ and rechecked for positivity, $R_{\mathrm{c}}>R_{\mathrm{t}}$, $R_{\mathrm{e}}>R_{\mathrm{t}}$, $0<L_{\mathrm{cyl}}<L_{\mathrm{c}}$, monotonic axial coordinates, and consistent injector-face, chamber, throat, and exit anchors.

\paragraph{Cooling search.}
The nominal GA used population 20, five generations, four elites, tournament size three, crossover probability 0.85, mutation probability 0.25, mutation scale 0.12, and two random immigrants. The bounded space contained 16--34 channels, inlet/outlet depths of 2--5/1--3.5~mm, widths of 1--1.8~mm, hot-wall thickness of 1.2--2.0~mm, and channel angle of $0$--$30^{\circ}$, with coolant-flow factor one. Dimensions were quantized to $0.1~\mathrm{mm}$, and channel widths or depths below $1.0~\mathrm{mm}$ were rejected. The prefilter also checked rib width, aspect ratio, and width-to-pitch compatibility.

RPA NozzleCooling used \texttt{addChannelWallSegment} with 75 stations, the Ievlev correlation, radiation emissivity 0.8, and no film cooling. A radiation-only fallback was rejected. Grid, Latin-hypercube, and feasibility-oriented candidate generators were also available; GA attribution required the corresponding recorded search mode and evolutionary settings. Run records separated invalid/pre-solver candidates, completed evaluations, hard-feasible candidates, Pareto-admissible candidates, and nondominated fronts. Conservative robustness criteria were evaluated separately from hard admission. The selected artifact retained the nozzle-exit coolant-inlet plan and stationwise structural-screen outputs.

\paragraph{Injector domain.}
The ten geometry or derived GPR features were $R_{\mathrm{Ox,out}}=1.5$--$1.7~\mathrm{mm}$, $R_{\mathrm{Fuel,out}}=2.8$--$3.1~\mathrm{mm}$, $L_{\mathrm{Fuel,out}}=4.4~\mathrm{mm}$, $L_{\mathrm{Fuel,out}}/(2R_{\mathrm{Fuel,out}})=0.7097$--$0.7857$, $L_{\mathrm{Ox,inner,out}}=10.5$--$13.5~\mathrm{mm}$, swirl angle $80$--$89.9^{\circ}$, swirl-pitch angle $0$--$20^{\circ}$, outlet-divergent angle $30$--$42^{\circ}$, $R_{\mathrm{Ox,swirl}}=0.7$--$1.1~\mathrm{mm}$, and $R_{\mathrm{Fuel,swirl}}=0.5$--$0.9~\mathrm{mm}$. Fixed original-propellant flow features covered 0.025--0.040~kg/s for LOX and 0.025--0.050~kg/s for RP-1. Numerical feature-domain validity was checked after count-specific flow allocation. Training-data water substitution used reference densities of 1145, 791.3, and 1000~kg/m$^3$ for LOX, RP-1, and water, respectively. The fixed geometry context was $w_{\mathrm{inj}}=0.75~\mathrm{mm}$, $R_{\mathrm{Ox,in}}=4.0~\mathrm{mm}$, and $R_{\mathrm{Fuel,in}}=5.0~\mathrm{mm}$. Search outputs were quantized to $0.1~\mathrm{mm}$ and $0.1^{\circ}$ before fresh prediction and joint screening.

\paragraph{Injector screening criteria.}
After quantization and fresh GPR evaluation, the screening procedure separates hard eligibility checks from optimization preferences. All twelve numerical features must remain within their declared bounds with extrapolation disabled, and the original-propellant single-element flows remain fixed at their count-specific allocations. Positive dimensions, the coupled radius and length-ratio constraints given below, and the stated angular limits are required; searched lengths and angles are quantized to $0.1$~mm and $0.1^{\circ}$ before fresh evaluation. The mean pressure drops must satisfy $\mu_{\Delta p,\mathrm{Fuel}}>200$~kPa and $\mu_{\Delta p,\mathrm{Ox}}>200$~kPa, corresponding to $0.20P_{\mathrm c}$ at $P_{\mathrm c}=1.0$~MPa. The provisional $84^{\circ}$--$92^{\circ}$ interval applies to the GPR-predicted mean spray angle during screening, including both endpoints; it was an operative window rather than a formally validated knowledge-base range and is not a hard interval on the VOF-computed angle. The optimization targets are a spray angle of $88^{\circ}$, Fuel pressure drop of 250~kPa, and Ox pressure drop of 230~kPa, which are preferences rather than equality constraints. Among candidates passing the preceding checks, the representative selection additionally requires $\mu_{\Delta p,j}-2\sigma_{\Delta p,j}>200$~kPa for both circuits, then minimizes $|\mu_{\theta}-88^{\circ}|$ and maximizes the smaller conservative pressure-drop margin. Predictive standard deviations and empirical coverage are retained for uncertainty reporting, but the $\mu-2\sigma$ values are not treated as calibrated confidence bounds. The four-element configuration matches the model package's element-count contract; the five-element search is an exploratory extension evaluated at its own single-element flows within the numerical feature bounds. These checks precede the independent configuration-level verification sequence.

For the fixed geometry context above, the coupled checks require $R_{\mathrm{Fuel,out}}-R_{\mathrm{Ox,out}}\geq1.2$~mm, $R_{\mathrm{Fuel,in}}-R_{\mathrm{Fuel,out}}\geq1.9$~mm, $R_{\mathrm{Ox,in}}-R_{\mathrm{Fuel,out}}\geq0.9$~mm, $R_{\mathrm{Fuel,in}}-R_{\mathrm{Ox,in}}\geq1.0$~mm, and $R_{\mathrm{Fuel,swirl}}\leq(R_{\mathrm{Fuel,in}}-R_{\mathrm{Fuel,out}})/2$, together with ordered, nonintersecting radii and a consistent $L_{\mathrm{Fuel,out}}/(2R_{\mathrm{Fuel,out}})$. The fixed $w_{\mathrm{inj}}=0.75$~mm is not a quantized search variable.

The representative NSGA-II searches use $f_j=[(\mu_j-t_j)/R_j]^2+2(\sigma_j/R_j)^2$, where $t_j$ is the stated optimization target and $R_j$ is the training-response range stored with the representative-design model. This uncertainty penalty acts during optimization; the additional $\mu-2\sigma$ condition is applied when selecting among freshly evaluated quantized candidates.

\paragraph{Surrogate model configuration.}
\label{app:gpr-configuration}
A single trained surrogate is used for the held-out module validation, the 140 post-selection evaluations, the representative-design searches, and the endpoint GPR predictions. Its training/validation/test split contains 549/118/118 samples for each response. Feature statistics, response-specific fitted hyperparameters, and subset records are retained with the model artifact. The module tests assess predictive performance over their evaluation populations; Table~\ref{tab:vof-endpoint} separately compares both count-specific injector candidates with their endpoint VOF responses.

The surrogate fits a separate GPR to each of the three responses using an amplitude-scaled Mat\'ern-$5/2$ kernel with automatic relevance determination (one length scale per feature), plus a white-noise kernel. The twelve inputs are standardized as $z_q=(x_q-m_q)/s_q$ using the stored feature means $m_q$ and scales $s_q$. No nonlinear response transform is applied; internal GPR normalization uses the training-response mean and standard deviation, with predictions returned in physical units. Kernel hyperparameters are optimized with the L\'evy-flight Whale Optimization Algorithm using training seed 42; the diagonal regularization parameter is $10^{-8}$. Existing training/validation/test assignments are retained during dataset import. Additional assignments, when required, use deterministic maximin space filling in normalized feature space; there is no random split seed.

\paragraph{Prediction-error normalization.}
For response $j$, the held-out error measures are
\begin{equation}
\mathrm{RMSE}_j=\left[\frac{1}{n}\sum_{i=1}^{n}(y_{ij}-\mu_{ij})^2\right]^{1/2},\qquad
\mathrm{NRMSE}_j=100\,\frac{\mathrm{RMSE}_j}{y_{j,\max}^{\mathrm{train}}-y_{j,\min}^{\mathrm{train}}}\%.
\label{eq:gpr-nrmse}
\end{equation}
Here, $n$ is the number of held-out samples and the denominator is the corresponding training-response range. For the reported module-validation NRMSE, those ranges are $113.128451^{\circ}$ for spray angle, 2744.383269~kPa for Fuel pressure drop, and 395.382108~kPa for Ox pressure drop. The training-range normalization differs from the endpoint relative deviations, which use the corresponding VOF window means as denominators. It also differs from the target-normalized score used to summarize the module-validation knee point.

\paragraph{Geometry and artifact handoff.}
PicoGK used millimetres, applied configuration changes transactionally, and restored the external configuration after execution. Completed runs retained nonempty assembled-solid and combined-fluid STLs, the applied geometric mapping, and generation logs. Binary validity, topology, local dimensions, connectivity, and meshing suitability required separate validator records. In the nozzle-configuration test, the TOC, $15^{\circ}$ conical, and $30^{\circ}/10^{\circ}$ parabolic-bell profiles retained 18, 14, and 18 sparse points, with minimum axial spacings of 0.9, 1.1, and 1.1~mm, respectively. Their solid/fluid triangle counts were 1,589,804/395,028 for TOC, 1,539,824/383,764 for the cone, and 1,508,448/377,168 for the bell.

 \bibliographystyle{elsarticle-num} 
 \bibliography{GPU}






\end{document}